\documentclass[10pt, letterpaper]{article}
\usepackage[margin=2cm]{geometry}
\usepackage{changes}
 
\usepackage{amssymb}
\usepackage{amsmath}
\usepackage{graphicx}

\usepackage{mathtools}

\usepackage{cite}

\newcommand{\dd}{\mathrm{d}}
\newcommand{\dv}[2]{\frac{\dd #1}{\dd #2}}

\usepackage{color}

\usepackage[ruled,vlined]{algorithm2e}

\usepackage{makecell,multirow}

\usepackage{nicefrac}   
\newcommand{\nfrac}[2]{\nicefrac{#1}{#2}}
\usepackage[super,comma,sort&compress]  
   {natbib}

\usepackage{CJKutf8} 

\title{Avoiding subtraction and division of stochastic signals using normalizing flows: NFdeconvolve}
\author{Pedro Pessoa$^{1,2}$, Max Schweiger$^{1,2}$, Lance W.Q. Xu \begin{CJK*}{UTF8}{gbsn}(徐伟青)\end{CJK*}$^{1,2}$, \\ Tristan Manha$^{1,2}$, Ayush Saurabh$^{1,2}$, Julian Antolin Camarena$^{1,2}$, \\  Steve Press\'e$^{1,2,3}$ \\  
$^1$Center for Biological Physics, Arizona State University,
Tempe, AZ, USA\\
$^2$Department of Physics, Arizona State University,
Tempe, AZ, USA\\
$^3$School of Molecular Sciences, Arizona State University,
Tempe, AZ, USA}
\date{}

\begin{document}

\maketitle

\abstract{
Across the scientific realm, we find ourselves subtracting or dividing stochastic signals. For instance, consider a stochastic realization, $x$, generated from the addition or multiplication of two stochastic signals $a$ and $b$, namely $x=a+b$ or $x = ab$. 
For the $x=a+b$ example, $a$ can be fluorescence background and $b$ the signal of interest whose statistics are to be learned from the measured $x$. 
Similarly, when writing $x=ab$, $a$ can be thought of as the illumination intensity and $b$ the density of fluorescent molecules of interest. 
Yet dividing or subtracting stochastic signals amplifies noise, and we ask instead whether, using the statistics of $a$ and the measurement of $x$ as input, we can recover the statistics of $b$.  
Here, we show how normalizing flows can generate an approximation of the probability distribution over $b$, thereby avoiding subtraction or division altogether. This method is implemented in our software package, \emph{NFdeconvolve}, available on GitHub with a tutorial linked in the main text.  
}


\section{Introduction}

Across scientific applications, measurements often involve a stochastic measurement $x$ resulting from the addition or multiplication of two random variables, $a$ and $b$. 
That is, 
\begin{equation}\label{x}
    x = a + b \quad \text{or}  \quad x = ab \ .
\end{equation}
Typically, the distribution over one of the random variables, $a$, can be obtained through control experiments, while the primary interest lies in the statistics of the other random variable, $b$, termed the ``signal''.

Examples of this paradigm include:
\begin{itemize}
	\item Background subtraction. Here, the background, $a$, is assumed pre-calibrated from experiments, and its statistics at each pixel are assumed to be known. Our goal is then to obtain the statistics over $b$ (often fluorescence signal) above the background from the total measurement $x$ while avoiding a naive stochastic subtraction of signal \cite{Fazel24}. 
	\item Illumination intensity correction. Here, the intensity of fluorescently emitted light from an object in microscopy is directly proportional to the product of the illumination intensity ($a$) and the object's density ($b$). The non-uniform illumination may again be calibrated over each pixel \cite{Model01,Peng17,Smith15}, and our goal is to avoid naive division or subtraction of stochastic signals to obtain $b$.
\end{itemize}

Put differently, from observations of $x$, $\{x\} = \{x_1,x_2, \ldots x_N\}$, and the known distribution of $a$ how can we learn the distribution over $b$? 
Concretely here, we denote the distribution over which $a$ is drawn as $a \sim p_A(\cdot|\theta_A)$ where $p_A$ represents a family of probability distributions and $\theta_A$ a set of parameters calibrated from experiments. Here, the symbol $\sim$ means `sampled from' or `has a distribution of'. As a concrete example, if $a$ is a normally distributed random variable, $\theta_A$ has two components, the mean value $\mu_A$ and the variance $\sigma_A^2$. In such a case, we write $\theta_A = (\mu_A, \sigma_A^2)$ and $p_A(a|\theta_A) = \frac{1}{\sqrt{2\pi \sigma_A^2}} \exp\left[- \frac{(a - \mu_A)^2}{2\sigma_A^2}\right]$.

Bayesian statistics provides a means of using $x$, {whose set of multiple observations we denote as $\{x\}$}, and $p_A(\cdot|\theta_A)$ to determine the distribution of $b$, but requires the \emph{often unknown} functional form of the probability distribution over $b$ \cite{Sivia06,Presse23}. 

To avoid specification of the form over the distribution over $b$ for the case of addition of $a$ and $b$, some Fourier-based strategies may be invoked. However, these methods can produce results that are problematic from a probabilistic perspective, \emph{e.g.}, negative probability densities or introduce high-frequency components that are likely artifacts arising from approximating the distribution with finite data \cite{TorregrosaCorts23}. 
To address these challenges, various regularization schemes are used \cite{Barraza-Felix99,krishan11,Trong14}. 

Another approach is to try to find a general well-studied family of probability distributions that would be able to, under some condition, approximate any general distribution. While some empirical guesses may serve as a starting point for what that distribution family may look like \cite{Brody07,Munkhammar17,Sriperumbudur17,Pessoa21},  one practical scheme may be to use non-parametric models for the distribution of $b$, often in the form of mixture models with a theoretically infinite number of components \cite{Kilic21,Bryan22,Presse23, TorregrosaCorts23}. However, as we will show later, even these flexible models may fail to accurately capture the true underlying distribution, especially when working with small datasets.

Recent advances in neural networks have introduced powerful methods in approximating general probability distributions \cite{Sukys22}. In particular, normalizing flows provide a neural network-based framework capable of representing a wide variety of probability distributions \cite{Rezende15, Papamakarios17, Durkan19, Dockhorn20, Kobyzev21, Stimper23}. Normalizing flows operate by applying a sequence of smooth, invertible transformations to a simple base distribution (typically Gaussian), enabling the construction of a more complex target distribution while still allowing for the exact computation of its density. 
This approach is grounded in the universal approximation theorem \cite{Goodfellow16}, stating that a sufficiently large neural network can approximate any continuous function to arbitrary precision. This property allows us to learn the distribution over $b$ without assuming a specific functional form, offering a more flexible and data-driven approach toward modeling complex distributions.

\begin{figure}
    \centering
    \includegraphics[width=.9\linewidth]{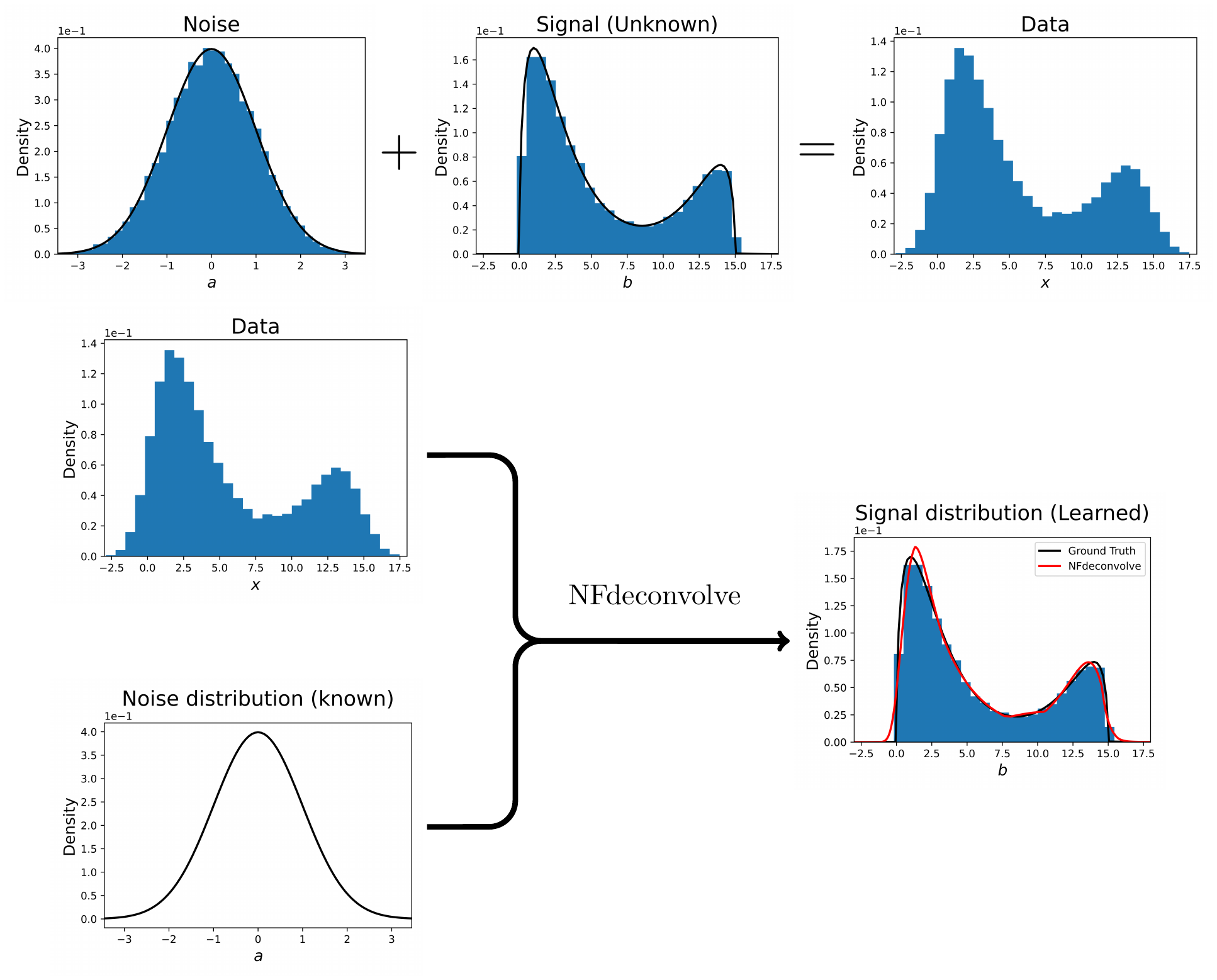}
    \caption{
    \textbf{Summary of \emph{NFdeconvolve}.}  
    The data formation process is assumed to involve two components: The noise, $a$, whose distribution is known and the signal, $b$, whose ground truth distribution is unknown. The observed data, $x$, is a convolution of these two components (in the example above, it is the sum $x = a + b$). \emph{NFdeconvolve} receives the combined data $\{x\}$, representing a set of observations of $x$ and the distribution of $a$ (but not the individual realizations of $a$'s) and produces an estimate of the distribution of $b$. For this example, whose detailed implementation can be seen in our GitHub \cite{github}, the resultant distribution of $b$ obtained by \emph{NFdeconvolve} is the one presented in red.
    }
    \label{fig:intro}
\end{figure}

Here, we present \emph{NFdeconvolve}, a software tool designed to learn the distribution of $b$ using normalizing flows. \emph{NFdeconvolve} is available in our GitHub repository \cite{github}. As demonstrated in the tutorial provided in the repository, users simply need to input their data \{x\} (\textit{i.e.}, the multiple observations of $x$) along with the probability distribution of $a$ -- given as an object drawn from the \emph{distributions} class of the PyTorch library \cite{pytorch} -- and outputs an estimate of the probability distribution over $b$. 
A cartoon illustration of how \emph{NFdeconvolve} operates is shown in Fig. \ref{fig:intro}.

In what follows, we explore two illustrative examples: one where $x$ is the sum of two random variables and another where $x$ is their product.
For each case, we will first employ Bayesian inference, using the same model that generated the simulation data, to obtain the probability distribution of $b$, leading to an optimal reconstruction. 
Following this, we will use a Bayesian framework with a mixture of Gaussian models, allowing us to observe the inherent challenges and limitations of using inaccurate models. Finally, we will present the results obtained using \emph{NFdeconvolve} to derive the distribution over $b$. Our analysis will evaluate the performance of these methods based on the data quality ({\it i.e.}, the extent to which $b$ contributes to the observed data) and quantity. We will demonstrate that normalizing flows provide a more reliable and accurate reconstruction when compared to this Gaussian mixture model, as measured by the Kullback-Leibler (KL) divergence between the distribution generating the simulated data (ground truth) and the probability distribution obtained by each method.

\section{Methods}\label{sec:methods}
The problem statement is verbally summarized as follows: we wish to learn the probability distribution over $b$ from a set of measurements $\{x\}$. Each element in this set is a deterministic function, a sum or a multiplication, of a realization of $b$ and the realization of another random variable $a$, whose distribution is known (by fitting realizations of $a$ independently). 

In the following subsection, we establish the necessary notation, and in Sec.~\ref{sec:solutions}, we describe three possible solution methods.

\subsection{Problem statement}\label{sec:model}
For the scope of the present article, we assume that each point $x_n \in \{x\}$ is independently and identically distributed from a sampling process summarized in the following  equations:
\begin{subequations} \label{model}
\begin{align}
    \theta_B &\sim p(\theta_B)     \ ,   \\
    a_n &\sim  p_A(\cdot|\theta_A) \ , \label{sample_a}   \\
    b_n|\theta_B &\sim  p_B(\cdot|\theta_B) \ ,  \label{sample_b} \\
    x_n  &= f(a_n,b_n)             \ ,   \label{makex}   
\end{align} 
\end{subequations}
with $p(\theta_B)$ being the prior for $\theta_B$ and $f$ is a deterministic function (normally addition or multiplication).
A point to clarify notation: probabilities with subscripts, such as $p_A$, are distributions for $a$ which belong to a specific family by construction.

In this setup, $a_n$ and $b_n$ are sampled independently. This independence allows us to express the probability distribution of $x$, with realizations $x_n$ as
\begin{equation} \label{px}
    p_X(x|\theta_A, \theta_B) = \int \dd b \ \dd a \ p_A(a|\theta_A) \ p_B(b|\theta_B) \ \delta(f(a, b) - x) \ .
\end{equation}
Using examples of addition and multiplication for $f$, we can simplify this expression. For addition,
\begin{equation}\label{sum-convolution}
    x_n = a_n + b_n \Rightarrow \quad p_X(x|\theta_A, \theta_B) = \int \dd b \ p_A(x - b|\theta_A) \ p_B(b|\theta_B) \ ,
\end{equation}
which is the convolution of $p_A$ and $p_B$. For multiplication,
\begin{equation}\label{prod-convolution}
    x_n = a_n b_n \Rightarrow \quad p_X(x|\theta_A, \theta_B) = \int \dd b \ p_A\left(\frac{x}{b}|\theta_A\right) \ p_B(b|\theta_B) \frac{1}{|b|}  \ ,
\end{equation}
known as the Mellin convolution of $p_A$ and $p_B$. If an explicit form for the integral in \eqref{px} is available, the sampling process in \eqref{model} can be simplified to $x_n \sim p_X(\cdot|\theta_A, \theta_B)$, replacing equations (\ref{sample_a}--\ref{makex}).

Since obtaining the distribution of $x$ from $b$ involves convolution, we will refer to learning the distribution of $b$ from $x$ as finding the deconvolved distribution.

\subsection{Theoretical basis for the methods}\label{sec:solutions}
This section introduces the mathematical theory for each of the three methods utilized in this paper, summarizing the methods and the associated internal parameters as given in Table \ref{tab:methods}. Here, we will continue to use the general form for $p_A$ and $p_B$, which were introduced in the previous section.

\begin{table}[]
\centering
\renewcommand{\arraystretch}{1.3} 
\footnotesize
\begin{tabular}{p{2.5cm}|p{5cm}|p{2.25cm}|p{3cm}|c}
& \multicolumn{1}{c|}{Learned Parameters} &  \multicolumn{2}{c|}{Deconvolved Distribution} & Definition
\\ \hline
Bayesian with known model & $\theta_B$, parameters of the underlying distribution & Reconstruction & $p(b|\{x\},\theta_A)$ & \eqref{Bayes-Reconstruction} \\ \cline{3-5}
                                           &                                                                       & MAP & $p_B(b|\theta_B^{\text{MAP}}(\{x\},\theta_A))$ & \eqref{Bayes-MAP} \\ \hline
Bayesian with Gaussian mixture & $\Psi_B$, encompassing the means $\{\mu_i\}$, variances $\{\sigma_i^2\}$, and weights $\{\rho_i\}$ of a Gaussian mixture & Reconstruction & $p(b|\{x\},\theta_A)$ &   \eqref{Gaussmix-reconstruction}\\ \cline{3-5}
                                               &                                                                       & MAP & $p_B(b|\Psi_B^{\text{MAP}}(\{x\},\theta_A))$ & \eqref{Gaussmix-MAP}\\ \hline
\emph{NFdeconvolve}                                  & $\phi$, internal parameters of the neural network & Reconstruction & $p_\text{NF}(b|\phi(\{x\},\theta_A))$ & \eqref{NF} \\ \hline
\end{tabular}
\caption{ \textbf{Summary of methods for obtaining the deconvolved distribution of $b$}. For each approach, the table shows the internal parameters, the corresponding deconvolved distributions, and references to the equations in which those were defined.}\label{tab:methods}
\end{table}

\subsubsection{Bayesian inference with known model for $b$}
Following the model established in Sec.~\ref{sec:model}, and assuming a functional form for $p_B$, by manipulating the rules of probability theory (\emph{i.e.}, using Bayes' theorem)  we learn $\theta_B$ as
\begin{equation} \label{Bayes}
    p(\theta_B|\{x\},\theta_A) \propto p(\{x\}|\theta_A,\theta_B) p(\theta_B)
\end{equation}
where
\begin{equation}
     p(\{x\}|\theta_A,\theta_B)  = \prod_{n=1}^N p_X(x_n|\theta_A,\theta_B)  \ .
\end{equation}
Here each factor $p_X(x_n|\theta_A,\theta_B)$ is computed from \eqref{px}.
We refer to $p(\{x\}|\theta_A,\theta_B)$ as the likelihood of the dataset and $p(\theta_B|\{x\},\theta_A)$ as the posterior over $\theta_B$.

While this Bayesian formalism provides a method to learn the parameters, $\theta_B$, from data, our goal is to obtain the distribution of $b$. A direct answer can be obtained by writing the joint probability of $b$ and $\theta_B$ and marginalizing over $\theta_B$, thus obtaining
\begin{equation} \label{Bayes-Reconstruction}
    p(b|\{x\},\theta_A) = \int \dd \theta_B \ p(b,\theta_B|\{x\},\theta_A) = \int \dd \theta_B \ p_B(b|\theta_B) \ p(\theta_B|\{x\},\theta_A) \ . 
\end{equation} 
In other words, integrating the functional form of $p_B$ weighted by the posterior, $ p(\theta_B|\{x\},\theta_A)$.

We refer to the distribution obtained by \eqref{Bayes-Reconstruction} as the reconstruction. As, in practice, the integral in \eqref{Bayes-Reconstruction} can rarely be calculated in closed form, we typically use sampling methods to generate values of $\theta_B$, which we then use to calculate a Monte Carlo approximation of \eqref{Bayes-Reconstruction} \cite{Presse23}. We provide details on how the Monte Carlo integration is performed for the examples discussed in our results section  in Appendix \ref{ApSec:MC}.

Besides the challenges in calculating the integral in  \eqref{Bayes-Reconstruction}, we highlight that $p(b|\{x\},\theta_A)$ in \eqref{Bayes-Reconstruction} is generally not a distribution in the functional form $p_B$. In other words, in general there is no $\theta_B^\ast$ such that $p(b|\{x\},\theta_A) = p_B(b|\theta_B^\ast)$.

For this reason, an alternative approach is to search for the value of $\theta_B$ that maximizes the posterior distribution, known as the \emph{maximum a posteriori} (MAP) estimate for the dataset ${x}$. The resulting distribution of $b$ is 
\begin{equation}\label{Bayes-MAP} 
p_B(b|\theta_B^\text{MAP} (\{x\},\theta_A) ) \ , \quad \text{where} \quad \theta_B^\text{MAP} (\{x\},\theta_A) = \arg \max\limits_{\theta_B} p(\theta_B|\{x\},\theta_A) \ , 
\end{equation} 
which follows the functional form $p_B$. For the remainder of this text, we will consider both the reconstruction in \eqref{Bayes-Reconstruction} and the MAP distribution in \eqref{Bayes-MAP} as valid methods for obtaining the deconvolved distribution. In particular, for large datasets, the posterior distribution should concentrate around the MAP estimate, $p(\theta_B|\{x\},\theta_A) \approx \delta ( \theta_B - \theta_B^\text{MAP} (\{x\},\theta_A)) $, making the two methods equivalent.

Although \eqref{Bayes-Reconstruction} and \eqref{Bayes-MAP} provide a mathematically consistent solution for finding the deconvolved distribution, they both assume that $p_B(b|\theta_B)$ is known. In most scientific endeavors, the model for $p_B$ is usually selected by studying the physical process behind the data generation. But what can be done without assuming that? We will discuss two possible methods in the following subsections.

\subsubsection{Bayesian with a Gaussian Mixture}\label{sec:gaussian_mixture}
To circumvent the need to specify the correct model for the deconvolved distribution, one natural approach is to opt for a general model that is well-studied, with the hope that such a model will adapt itself to be sufficiently close to the underlying model. One popular choice is to use a non-parametric mixture of Gaussians\cite{Moulines97,Santamaria99,Bovy11,TorregrosaCorts23}, as it is versatile enough to capture a wide range of underlying distributions. 

Here, we represent the parameters of this mixture as $\Psi_B = \{ \bar{\mu}, \bar{\sigma^2}, \bar{\rho} \}$, encompassing the means $\bar{\mu} = (\mu_1, \mu_2, \mu_3, \ldots)$, variances $\bar{\sigma^2} = (\sigma^2_1, \sigma^2_2, \sigma^2_3, \ldots)$, and weights $\bar{\rho} = (\rho_1, \rho_2, \rho_3, \ldots)$. We then assume a probability for $b$ of the form
\begin{equation}\label{infGaussian}
    p_B(b|\Psi_B) = \sum_{i=1}^\infty \rho_i  \frac{1}{\sqrt{2\pi \sigma_i^2}} \exp\left[- \frac{(b-\mu_i)^2}{2 \sigma^2_i} \right] \ . 
\end{equation}
Thus, $\Psi_B$ assumes a role similar to $\theta_B$ in Bayesian inference with a known model, though we do not assume the Gaussian mixture model to \emph{necessarily} be the underlying distribution generating the data. 

It is important to mention that, in the context of approximating an unknown distribution, the mixture components generally do not have an intrinsic physical meaning, the Gaussian mixture serves as a  parametrization of the space of probability distributions. 
However, in certain applications, such as separating signals from distinct underlying sources, the components can acquire a meaningful interpretation -- \emph{e.g.}, when the data arise from multiple species, each characterized by its own emission model. In such cases, the mixture weights correspond to the relative proportions of these contributing species \cite{Bryan22,Sorek24,Sgouralis24,Teng24}

Though the non-parametric mixture of Gaussians model in \eqref{infGaussian} theoretically involves an infinite number of parameters, for computational feasibility, in all results presented, we restrict the maximum number of components $i$ to 20. 
This approach requires a large dataset to avoid overfitting whilst accommodating functional forms for the distribution over $b$ that are not readily accommodated by a small number of Gaussians. 
We can calculate the likelihood of a data point, $p_X(x|\theta_A, \Psi_B)$, by replacing $\theta_B$ with $\Psi_B$ in \eqref{px}, from which follows the posterior for $\Psi_B$, $p(\Psi_B|\{x\},\theta_A)$ as :
\begin{equation} \label{BayesGM}
    p(\Psi_B|\{x\},\theta_A) \propto p(\{x\}|\theta_A,\Psi_B) p(\Psi_B) \ .
\end{equation}
The prior distribution, $p(\Psi_B)$, is especially critical for non-parametric models to address the degeneracy caused by permuting the component labels $i$ in \eqref{infGaussian}. The specific prior used in this work is detailed in Appendix \ref{ApSec:GMprior}. 

From this, we obtain the equivalent reconstruction as
\begin{equation} \label{Gaussmix-reconstruction}
    p(b|\{x\},\theta_A) = \int \dd \Psi_B \ p(b,\Psi_B|\{x\},\theta_A) = \int \dd \Psi_B \ p_B(b|\Psi_B) \ p(\Psi_B|\{x\},\theta_A) \ ,
\end{equation} 
and the equivalent MAP distribution as 
\begin{equation}\label{Gaussmix-MAP} 
p_B(b|\Psi_B^\text{MAP} (\{x\},\theta_A) ) \ , \quad \text{where} \quad \Psi_B^\text{MAP} (\{x\},\theta_A) = \arg \max\limits_{\Psi_B} p(\Psi_B|\{x\},\theta_A) \ . 
\end{equation} 
As in Bayesian inference with a known model, we use samples of $\Psi_B$ from the posterior and calculate a Monte Carlo approximation of \eqref{Gaussmix-reconstruction}. Further details can be found in our GitHub repository \cite{github}.

\subsubsection{Normalizing Flows and \emph{NFdeconvolve}}\label{sec:nfdeconvolve}
While mixtures of Gaussians are popular in obtaining a probability distribution of an unknown form, another neural-network-based framework for reconstructing probability distributions has recently gained  traction: normalizing flows \cite{Rezende15, Papamakarios17, Durkan19, Dockhorn20, Kobyzev21, Stimper23}. 
Normalizing flows offer flexibility and expressiveness by leveraging neural networks to model complex distributions through a sequence of transformations. 
The method presented in this paper, called \emph{NFdeconvolve}, uses normalizing flows to retrieve the deconvolved distribution.

The fundamental idea behind normalizing flows is to start with a simple and well-understood base distribution, typically Gaussian, $p_Z$ for a variable $z$. This base distribution is then transformed into the target distribution using a series of invertible and smooth mappings. Specifically, we define a function $f_\phi(z)$ parameterized by a neural network, where $\phi$ represents all of the network's internal parameters. The neural network architecture is selected such that $f_\phi$ is invertible and smooth. In \emph{NFdeconvolve}, we employed the architecture known as neural spline flow \cite{Durkan19} found in the \emph{normflows} package \cite{Stimper23}.

The probability distribution of the image of $z$ through $f_\phi$ is given by 
\begin{equation}
    p(f_\phi(z)) = p_Z(z) \left|\dv{f_\phi}{z}\right|^{-1} \ .
\end{equation}
We then define the probability distribution generated by the normalizing flows with $f_\phi$, which we refer to as $p_\text{NF}$ as the probability of the image of $z$  
\begin{equation}\label{p_nf-def}
    p_\text{NF}(b|\phi) = p_Z\left(f^{-1}_\phi(b)\right) \left|\dv{f_\phi}{z} \left(f^{-1}_\phi(b)\right) \right|^{-1} \ ,
\end{equation} 
as a possible distribution of $b$ parametrized by the neural network parameters $\phi$. Since $f_\phi$ is invertible and smooth, both $f^{-1}_\phi(b)$ and $\dv{f_\phi}{z}$ are well-defined, thus ensuring easy calculation of $p_\text{NF}(b|\phi)$.

When using normalizing flows to approximate the unknown distribution of $b$, we maximize the likelihood of the observed data under $p_{NF}$. In other words, if we have direct realizations of $b$, $\{b\} = \{b_1, b_2, \ldots, b_N\}$, we adjust the parameters $\phi$  such that  the total likelihood of  $\{b\}$, $\prod\limits_{n=1}^N p_\text{NF}(b_n|\phi)$, is maximized. If we treat the observations as samples from the target distribution, this is equivalent to minimizing the approximate KL divergence between the target distribution and the distribution obtained from the normalizing flows \cite{Rezende15}.

However, for the problem at hand -- finding the deconvolved distribution of $b$ from observations of the composite variable $x$ -- we instead train the network to maximize the likelihood of $x$ arising from the distribution of $b$ given by $p_\text{NF}$.  
Consistently with \eqref{px} we define
\begin{equation}\label{cov-nf}
    p(x|\phi,\theta_A) = \int \dd b \ \dd a \ p_A(a|\theta_A) \ p_\text{NF}(b|\phi) \ \delta (f(a,b)-x)\ .
\end{equation}
and we say that we find the deconvolved distribution of $b$ as the one obtained by the neural network parameters, $\phi$, that maximizes the likelihood of $\{x\}$ (in other words, train the neural network with the negative of the likelihood's logarithm as the loss function). That means the resulting distribution of $b$ is now
\begin{equation}\label{NF} 
p_\text{NF}(b|\phi^\ast(\{x\},\theta_A) ) \ , \quad \text{where} \quad \phi^\ast (\{x\},\theta_A) = \arg \max\limits_{\phi} \prod_{n=1}^N p(x_n|\phi,\theta_A) \ , 
\end{equation}
with the factors within the product given by \eqref{cov-nf}. Note that this is equivalent to a MAP distribution from a Bayesian formalism where one assumes the prior distribution for the network parameters, $\phi$, to be uniform.

We refer to our GitHub repository \cite{github} for details on how to calculate the integral in \eqref{cov-nf} and how we use the optimization libraries within PyTorch \cite{pytorch}. 

\begin{figure}
    \centering
    \includegraphics[width=.49\linewidth]{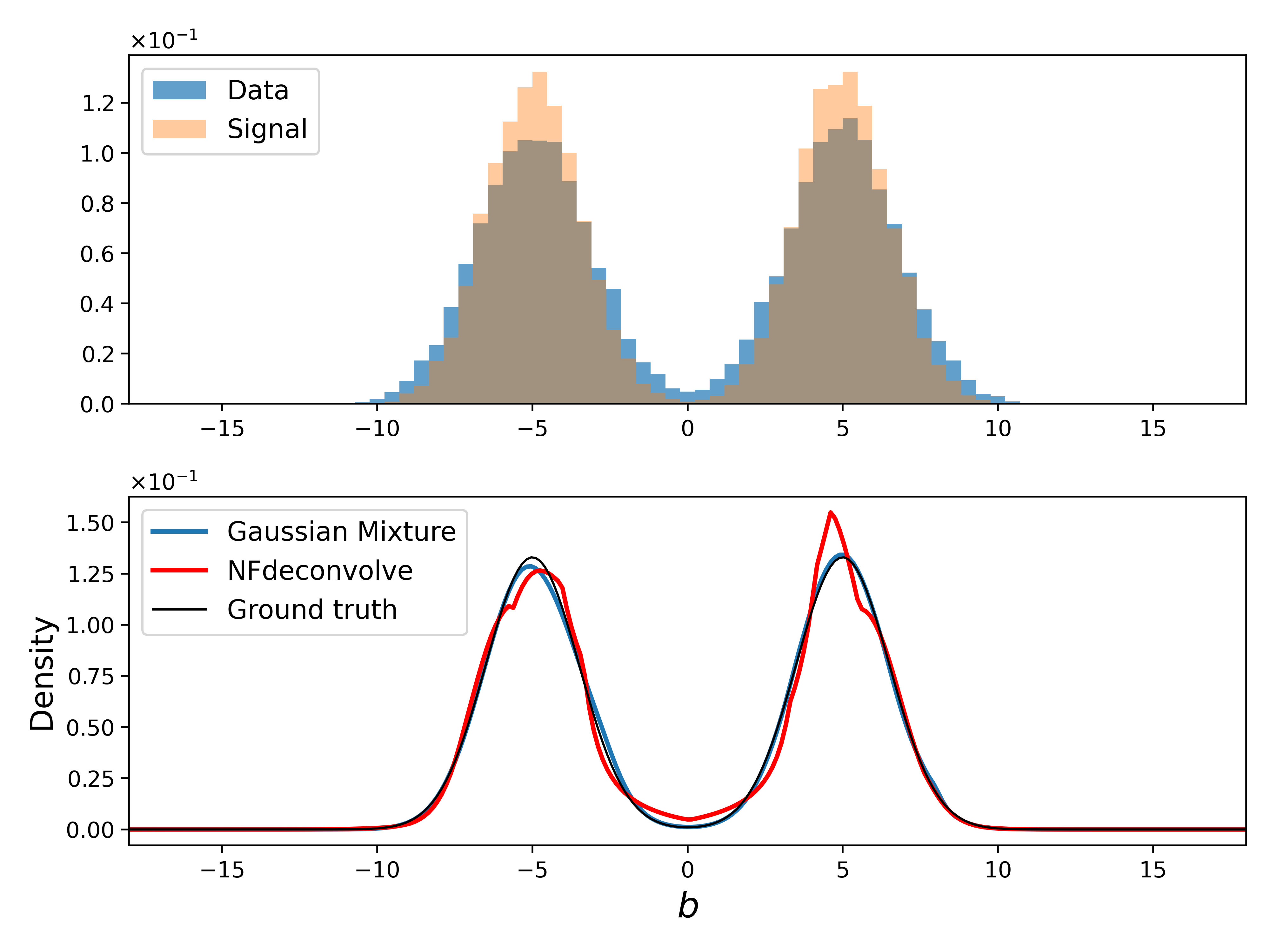}
    \includegraphics[width=.49\linewidth]{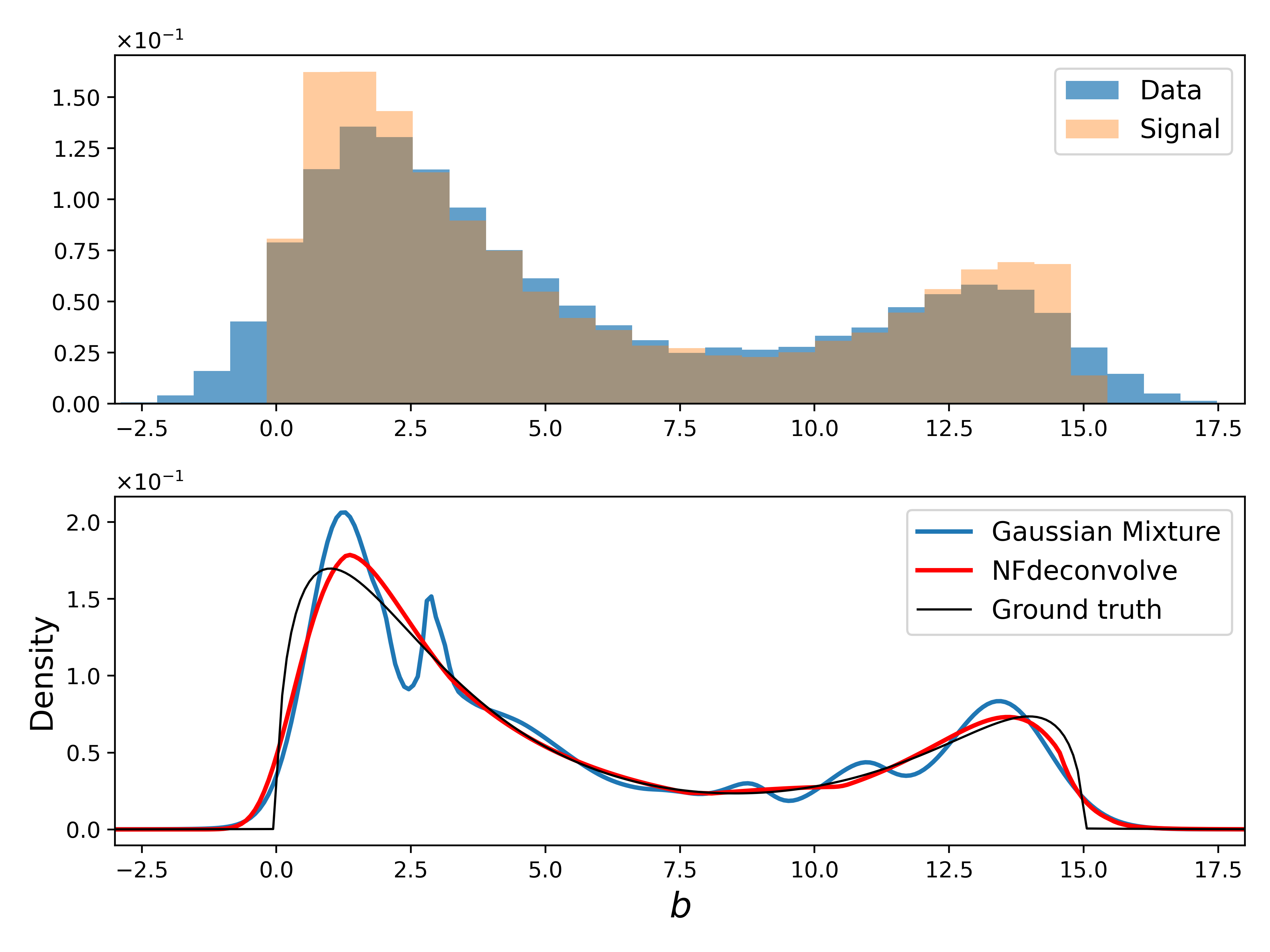}
    \caption{
    \textbf{Comparison of \emph{NFdeconvolve} and a Gaussian mixture model for obtaining the deconvolved distribution in two scenarios.} 
    The top row shows the true signal distribution $b$ and the observed data $x$ in two scenarios: on the left, $b$ is generated from a mixture of Gaussians; on the right, from a mixture of Gamma and inverted Gamma distributions.  The bottom row presents the deconvolved distributions obtained using both a Bayesian approach with a Gaussian mixture model and \emph{NFdeconvolve}. On the left, where the data matches the Gaussian mixture assumption, both methods perform well, with the Gaussian mixture model achieving a more precise match. However, on the right, where the data distribution does not align with the Gaussian model, \emph{NFdeconvolve} significantly outperforms the Gaussian mixture approach. Further quantitative comparisons will be presented  Sec.~\ref{sec:results}.}
    \label{fig:brute-comparison}
\end{figure}

To demonstrate the need for a method like \emph{NFdeconvolve}, Fig. \ref{fig:brute-comparison} compares its performance to a Bayesian with Gaussian mixture model, as described in Sec.~\ref{sec:gaussian_mixture}. 
While the Gaussian mixture performs well when the underlying distribution of $b$ is, indeed, a Gaussian mixture, it fails under model mismatch with non-Gaussian where \emph{NFdeconvolve}, in contrast, provides a more accurate deconvolved distribution.

\section{Results}\label{sec:results}
Having discussed the problem in Sec.~\ref{sec:model} and possible solution methods in Sec.~\ref{sec:solutions}, we now provide examples of their usage.  
We will first present an example where the signal is a sum of the two stochastic variables. $x = a+b$ with $a$ being samples from a Gaussian while $b$ is sampled from a Gamma distribution and another where the data is the product of the two stochastic variables, $x = ab$, sampled from the same distributions.

For each example, we generate multiple synthetic datasets using different parameterizations for the ground truth distribution of $b$. These datasets are used to compute the signal-to-noise ratio (SNR), which serves as a measure of data quality; higher SNR values correspond to signals that are more ``visible'' or less influenced by the noise, as one can expect all methods to exhibit improved performance at higher SNR. Additionally, we test datasets with the same SNR but varying numbers of datapoints ($N$), as larger datasets generally allow methods to perform better.

To quantitatively compare the methods, we compute the KL divergence between the ground truth distribution and the distributions inferred by each method. The results demonstrate how each method performs across datasets with varying sizes ($N$) and data quality (SNR), as described in Sec.~\ref{sec:methods}.

\subsection{Sum of two random variables example}\label{sec:sum}

\begin{figure}
    \centering
    \includegraphics[width=\linewidth]{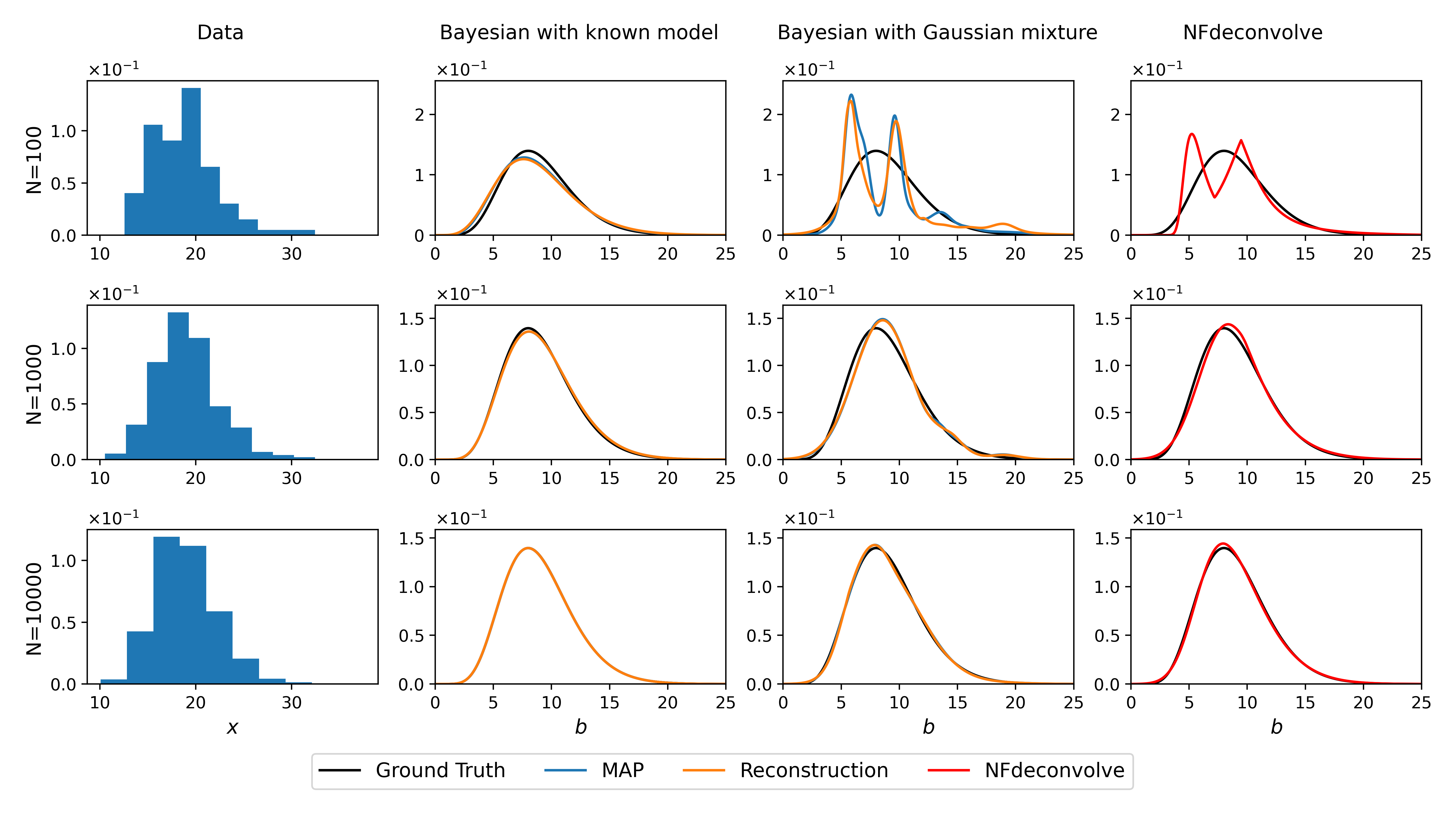} \vspace{-.6cm}
    \caption{\textbf{Distributions obtained by the solution methods for the sum of two random variables example.} Here $b$ is sampled from a ground truth Gamma distribution, as in \eqref{Gamma}, with parameters $\alpha_B = 9$ and $\beta_B = 1$, while $a$ is sampled from a Gaussian with mean $\mu_A = 10$ and variance $\sigma_A^2 = 1$. In each row, we change the number of data points used and show the distributions obtained by each method. In the two Bayesian methods, we show both the MAP and the reconstruction. As expected, the Bayesian method with the known model finds the correct distribution with fewer data points. For the other methods, we see that the Gaussian mixture presents some overfitting, represented by the ``peaks'' in the corresponding column, while the normalizing flows approach the ground truth distribution in a smoother way. Later, we will quantitatively confirm this result.}
    \label{fig:sum}
\end{figure}

For our first example, we use a Gaussian distribution for $p_A$ with a known mean $\mu_A$ and variance $\sigma_A^2$, $\theta_A = \{\mu_A, \sigma_A^2\}$. Here, we are supposed to recover $p_B$ assumed to be a Gamma distribution with parameters $\theta_B = \{\alpha_B, \beta_B\}$ 
\begin{equation}\label{Gamma}
    p_B(b|\theta_B) = \frac{\beta_B^{\alpha_B}}{\Gamma(\alpha_B) }b^{\alpha_B-1} e^{- \beta_B b} \ ,
\end{equation}
with $\Gamma$ representing the gamma function for $b>0$ and zero otherwise.  
To the best of our knowledge, the distribution for $x = a + b$ in this example does not have a closed-form expression. In Fig. \ref{fig:sum}, we show how each method performs for the same ground truth distribution for $b$ and different dataset sizes. The details of how the data was generated and the numerical schemes used to obtain each distribution can be found in our GitHub repository \cite{github}. We note that the system quickly identifies the correct distribution when the correct model is provided. However, we must rely on the other two methods when the correct model is not provided. As observed in Fig. \ref{fig:sum}, the Gaussian mixture tends to produce a ``rougher'' distribution that attempts to identify patterns in the likelihood of the finite data rather than capturing the overall trend. In contrast, the NFdeconvolve is smoother, even for small datasets.

We now proceed to demonstrate how these methods perform with datasets of different sizes and overall quality. Here, to represent data of varying quality, we define the SNR as the ratio of variances between the (ground truth) distribution of $b$ and the distribution of $a$:
\begin{equation}
    \text{SNR} \doteq  \frac{\sigma_b^2}{\sigma_a^2} \quad \text{with}\quad \sigma_b^2 \doteq \langle b^2\rangle - \langle b\rangle^2 \ .
\end{equation}
Note that the usage of $\sigma_b^2$ is consistent with our previous notation. Specifically, the definition of $\sigma_b^2$ aligns with the use of $\sigma_a^2$ as a variable because, for Gaussian distributions, the $\sigma_a^2$ parameter represents the variance. 
Also, the SNR does not change when $a$ or $b$ are shifted by summing a constant term. A small SNR means that the noise (or the stochasticity of the variable we are not interested in) is dominant, while a large SNR means that the stochasticity observed in the measurement $x$ is mostly generated by the stochasticity of the variable we are interested in, $b$. In the particular case we are treating here, where $b$ is Gamma distributed, the SNR can be simplified as $\text{SNR} = \frac{\alpha_B}{\beta_B ^2 \sigma_a^2}$.

To evaluate the performance of each solution method, we calculate the KL divergence from the obtained distribution to the ground truth distribution. This is expressed as:
\begin{equation}\label{KL}
    KL[q] = \int \dd b \  p(b|\theta_B^\text{GT}) \log \frac{p(b|\theta_B^\text{GT}) }{q(b)} \ , 
\end{equation}
where $\theta_B^\text{GT}$ represents the parameters of the ground truth distribution, and $q$ is the distribution being evaluated. 
For example, when evaluating the distribution obtained by the NFdeconvolve, $p_\text{NF}(b|\phi^*(\{x\},\theta_A))$ defined in \eqref{NF}, we calculate the KL divergence as with $q(b)= p_\text{NF}(b|\phi^*(\{x\},\theta_A)$. 

\begin{figure}
    \centering
    \includegraphics[width=.9\linewidth]{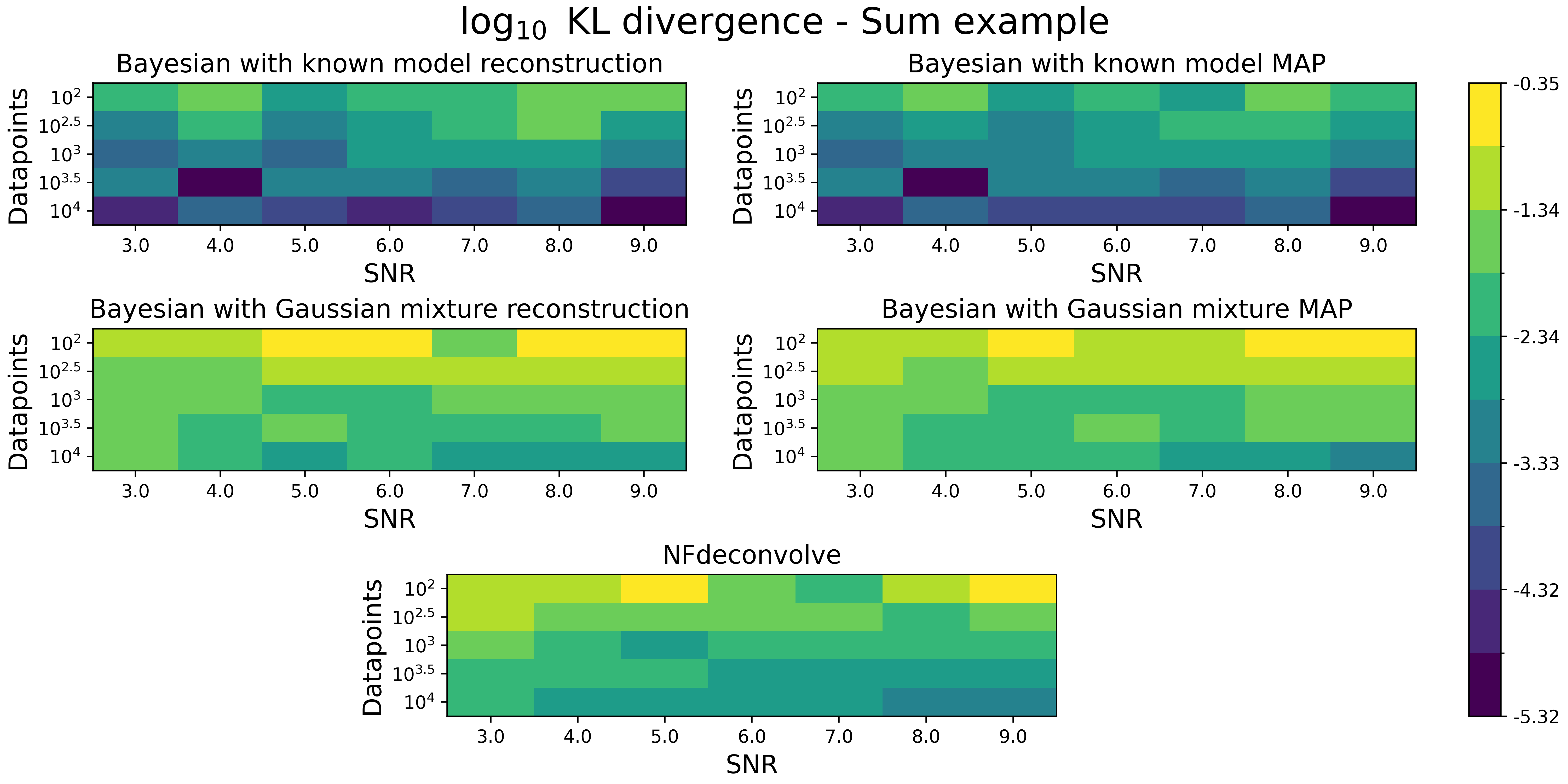}
    \caption{\textbf{Divergence between the ground truth distribution and the distributions obtained by each method in the sum of two random variables example measured by the logarithm of the KL divergence.} Each square within the figure was obtained with a synthetic dataset where $b$ is Gamma distributed with parameter $\beta_B = 1$ and the other parameter $\alpha_B$ is changed to generate datasets with different SNR. In all cases, $a$ is sampled from a Gaussian with mean $\mu_A = 10$ and variance $\sigma_A^2 = 1$. As expected, we see smaller KL divergence values (darker colors) for larger SNR and data sizes. The Bayesian method, along with the known model, is able to obtain a distribution much closer to the ground truth than all others. However, when the model is unknown, the normalizing flows generally obtain distributions with smaller KL divergence. This confirms our result in Fig. \ref{fig:sum} that normalizing flows are able to better approximate the unknown distribution than the Gaussian mixture by avoiding overfitting. }
    \label{fig:KLsum}
\end{figure}

We present the KL divergence across datasets of different sizes and SNR in Fig. \ref{fig:KLsum}. Here, we confirm the results presented in Fig. \ref{fig:sum}: while the Bayesian method with the correct model performs considerably better (represented by a KL divergence that is an order of magnitude smaller), among the Gaussian mixture and the normalizing flows results, we see that the latter performs significantly better. {In all cases, it is also possible to observe how, as expected, the KL divergence decreases, indicating that the distribution found by the method approaches the ground truth with more data and higher SNR.}

\subsection{Product of two random variables example}\label{sec:product}

Similarly to the previous example, we sample $a$ from a Gaussian distribution with $b$ from a Gamma distribution. The only difference is that the data is obtained as the product, $x=ab$.

Although the mathematics for obtaining the distribution of $x=ab$ can be performed by solving the integral in \eqref{prod-convolution}, we have found that it can be numerically unstable. As a consequence, we instead performed all necessary calculations in logarithmic space. Specifically, we use the identity $\log x = \log a + \log b$ to transform the product of random variables into the sum of random variables. This requires that we transform the probability distribution of $a$, $p_A$, into logarithmic space, which is straightforwardly obtained as $p(\log a|\theta_A) = a p(a|\theta_A)$. Further details with an application for each method can be found in Appendix \ref{ApSec:MC}.

\begin{figure}
    \centering
    \includegraphics[width=\linewidth]{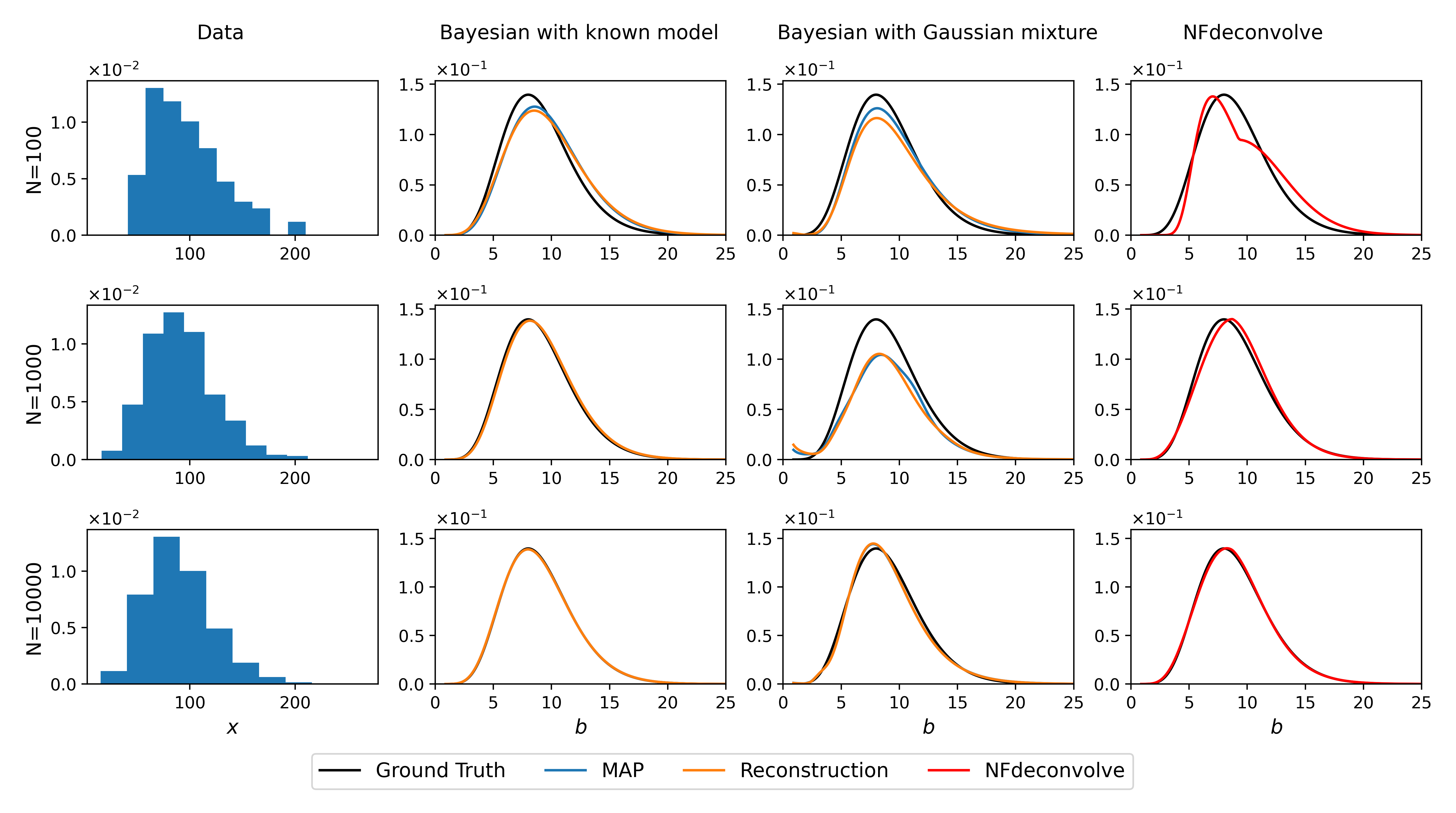}\vspace{-.6cm}
    \caption{\textbf{Distributions obtained by the solution methods for the product of two random variables example.} Here $b$ is sampled from a ground truth Gamma distribution, as in \eqref{Gamma}, with parameters $\alpha_B = 9$ and $\beta_B = 1$, while $a$ is sampled from a Gaussian with mean $\mu_A = 10$ and variance $\sigma_A^2 = 1$. In each row, we change the number of data points used and show the distributions obtained by each method. Consistently with Fig. \ref{fig:sum}, the Bayesian method with the known model finds the correct distribution with fewer data points. Similarly, among the other methods that do not require knowing the model, the normalizing flows method avoids the overfitting seen in the Gaussian mixture.  }
    \label{fig:prod}
\end{figure}

As with the previous example, Fig. \ref{fig:prod}, we show how each method performs for the same ground truth distribution for $b$ and different dataset sizes. Mirroring the previous results, {the system identifies the correct distribution when provided with the correct model, even with a small dataset}. However, without knowledge of the true model, the NFdeconvolve is considerably less prone to overfitting than the Gaussian mixture. Numerical details can be found in our GitHub repository \cite{github}.

We move to demonstrate how these methods perform with datasets of different sizes and overall quality, represented by SNR. However, while for the sum example, we defined the SNR as the ratio of variances between the (ground truth) distribution of $b$ and the variance of $a$, here in the product example, we change the definition to the ratio of relative variances, meaning 
\begin{equation}
    \text{SNR} \doteq  \frac{\left(\nfrac{\sigma_b}{\langle b \rangle}\right)^2} 
    {\left(\nfrac{\sigma_a}{\langle a \rangle}\right)^2} \ .
\end{equation}
Under this modified definition, when $a$ or $b$ are scaled by a constant factor, the SNR does not change. In the particular case we are treating here, the SNR can be simplified to the straightforward formula $\text{SNR} = \frac{\sigma_a^2}{\mu_a^2 \alpha_B}$.

\begin{figure}
    \centering
    \includegraphics[width=.9\linewidth]{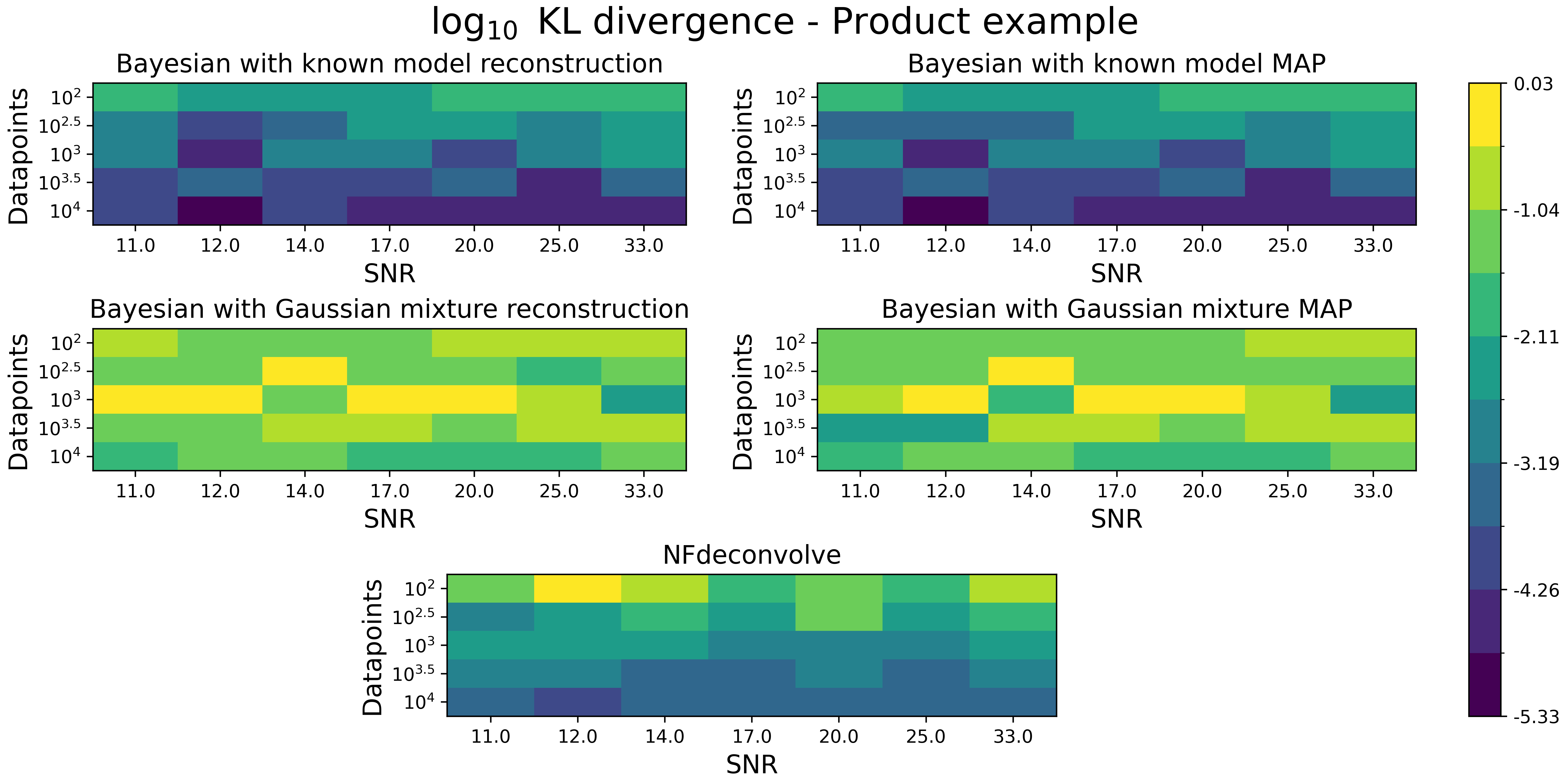}
    \caption{\textbf{Divergence between the ground truth distribution and the distributions obtained by each method in the product of two random variables example measured by the logarithm of the KL divergence.} Each square within the figure was obtained with a synthetic dataset where $b$ is Gamma distributed with parameter $\beta_B = 1$ and the other parameter $\alpha_B$ is changed to generate datasets with different SNR. In all cases, $a$ is sampled from a Gaussian with mean $\mu_A = 10$ and variance $\sigma_A^2 = 1$. Consistently with Fig. \ref{fig:KLsum}, we observe that the normalizing flows method obtains distributions with smaller KL divergence when compared to the Gaussian mixture.}
    \label{fig:KLprod}
\end{figure}

Finally, we present the KL divergence across datasets of different sizes and SNR in Fig. \ref{fig:KLprod}, confirming the results presented in Fig. \ref{fig:prod}. In accordance with the results for the sum example, the results obtained with \emph{NFdeconvolve} are significantly better than those obtained using the Gaussian mixture.

\section{Conclusion}

In this short article, we address the challenge of determining the distribution of a stochastic variable through indirect measurements. We model this as the composition of two stochastic variables, where the distribution of one is known, and the goal is to infer the distribution of the other using observations of the composed realizations. 

To rigorously tackle this issue, we delve into the Bayesian formalism required to solve the problem. However, this approach necessitates selecting a model, which is not always feasible. Without a known model, we propose two alternative methods: a Bayesian approach using a non-parametric mixture of Gaussians, and a method that approximates the distribution of interest through the neural network-based technique known as normalizing flows. The latter method is implemented as our software tool, \emph{NFdeconvolve}, which we have made publicly available in our GitHub repository, with tutorials for implementation. 

The examples shown in Figs. \ref{fig:sum} and \ref{fig:prod}  reveal that while neither method can fully replace the knowledge of the correct model, the distribution obtained via normalizing flows is significantly less prone to overfitting. This is supported by the quantitative analysis in Figs. \ref{fig:KLsum} and \ref{fig:KLprod}, which show that when applied to synthetic data of varying sizes and SNR, the distribution estimated by normalizing flows more closely aligns with the ground truth, achieving approximately smaller lower KL divergence to the ground truth distribution.

\section*{Acknowledgments}
SP acknowledges lively discussions with his colleagues Profs. Seungeun Oh and Suckjoon Jun that inspired this work in the context of Raman and support from the NIH (Grant No. R35GM148237) and NSF (Grant No. 2310610)

\bibliography{refs}

\begin{thebibliography}{32}
\providecommand{\natexlab}[1]{#1}
\providecommand{\url}[1]{\texttt{#1}}
\providecommand{\href}[2]{#2}
\providecommand{\path}[1]{#1}
\providecommand{\DOIprefix}{doi: }
\providecommand{\ArXivprefix}{arXiv: }
\providecommand{\URLprefix}{URL: }
\providecommand{\Pubmedprefix}{pmid: }
\providecommand{\doi}[1]{\href{http://dx.doi.org/#1}{\path{#1}}}
\providecommand{\Pubmed}[1]{\href{pmid:#1}{\path{#1}}}
\providecommand{\BIBand}{and}
\providecommand{\bibinfo}[2]{#2}
\ifx\xfnm\undefined \def\xfnm[#1]{\unskip,\space#1}\fi
\makeatletter\def\@biblabel#1{#1.}\makeatother
\bibitem[{Fazel et~al.(2024)Fazel, Grussmayer, Ferdman, Radenovic, Shechtman, Enderlein and Press\'e}]{Fazel24}
\bibinfo{author}{Fazel, M.}, \bibinfo{author}{Grussmayer, K.S.}, \bibinfo{author}{Ferdman, B.}, \bibinfo{author}{Radenovic, A.}, \bibinfo{author}{Shechtman, Y.}, \bibinfo{author}{Enderlein, J.}, and \bibinfo{author}{Press\'e, S.} (\bibinfo{year}{2024}). \bibinfo{title}{Fluorescence microscopy: A statistics-optics perspective}.
\newblock \bibinfo{journal}{Rev. Mod. Phys.} \emph{\bibinfo{volume}{96}}, \bibinfo{pages}{025003}. \DOIprefix\doi{10.1103/RevModPhys.96.025003}.
\bibitem[{Model and Burkhardt(2001)}]{Model01}
\bibinfo{author}{Model, M.A.}, and \bibinfo{author}{Burkhardt, J.K.} (\bibinfo{year}{2001}). \bibinfo{title}{A standard for calibration and shading correction of a fluorescence microscope}.
\newblock \bibinfo{journal}{Cytometry} \emph{\bibinfo{volume}{44}}, \bibinfo{pages}{309–316}. \DOIprefix\doi{{10.1002/1097-0320(20010801)44:4<309::aid-cyto1122>3.0.co;2-3}}.
\bibitem[{Peng et~al.(2017)Peng, Thorn, Schroeder, Wang, Theis, Marr and Navab}]{Peng17}
\bibinfo{author}{Peng, T.}, \bibinfo{author}{Thorn, K.}, \bibinfo{author}{Schroeder, T.}, \bibinfo{author}{Wang, L.}, \bibinfo{author}{Theis, F.J.}, \bibinfo{author}{Marr, C.}, and \bibinfo{author}{Navab, N.} (\bibinfo{year}{2017}). \bibinfo{title}{A basic tool for background and shading correction of optical microscopy images}.
\newblock \bibinfo{journal}{Nature Communications} \emph{\bibinfo{volume}{8}}, \bibinfo{pages}{14836}. \DOIprefix\doi{10.1038/ncomms14836}.
\bibitem[{Smith et~al.(2015)Smith, Li, Piccinini, Csucs, Balazs, Bevilacqua and Horvath}]{Smith15}
\bibinfo{author}{Smith, K.}, \bibinfo{author}{Li, Y.}, \bibinfo{author}{Piccinini, F.}, \bibinfo{author}{Csucs, G.}, \bibinfo{author}{Balazs, C.}, \bibinfo{author}{Bevilacqua, A.}, and \bibinfo{author}{Horvath, P.} (\bibinfo{year}{2015}). \bibinfo{title}{{CIDRE: an illumination-correction method for optical microscopy}}.
\newblock \bibinfo{journal}{Nature Methods} \emph{\bibinfo{volume}{12}}, \bibinfo{pages}{404–406}. \DOIprefix\doi{10.1038/nmeth.3323}.
\bibitem[{Sivia and Skilling(2006)}]{Sivia06}
\bibinfo{author}{Sivia, D.}, and \bibinfo{author}{Skilling, J.} (\bibinfo{year}{2006}). \bibinfo{title}{Data Analysis: A Bayesian Tutorial}.
\newblock Oxford science publications. \bibinfo{publisher}{OUP Oxford}.
\newblock ISBN \bibinfo{isbn}{9780198568322}.
\bibitem[{Press{\'e} and Sgouralis(2023)}]{Presse23}
\bibinfo{author}{Press{\'e}, S.}, and \bibinfo{author}{Sgouralis, I.} (\bibinfo{year}{2023}). \bibinfo{title}{Data Modeling for the Sciences: Applications, Basics, Computations}. \bibinfo{publisher}{Cambridge University Press}.
\newblock ISBN \bibinfo{isbn}{9781009098502}.
\bibitem[{Torregrosa-Cortés et~al.(2023)Torregrosa-Cortés, Oriola, Trivedi and Garcia-Ojalvo}]{TorregrosaCorts23}
\bibinfo{author}{Torregrosa-Cortés, G.}, \bibinfo{author}{Oriola, D.}, \bibinfo{author}{Trivedi, V.}, and \bibinfo{author}{Garcia-Ojalvo, J.} (\bibinfo{year}{2023}). \bibinfo{title}{{Single-cell Bayesian deconvolution}}.
\newblock \bibinfo{journal}{iScience} \emph{\bibinfo{volume}{26}}, \bibinfo{pages}{107941}. \DOIprefix\doi{10.1016/j.isci.2023.107941}.
\bibitem[{Barraza-Felix and Frieden(1999)}]{Barraza-Felix99}
\bibinfo{author}{Barraza-Felix, S.}, and \bibinfo{author}{Frieden, B.R.} (\bibinfo{year}{1999}). \bibinfo{title}{Regularization of the image division approach to blind deconvolution}.
\newblock \bibinfo{journal}{Appl. Opt.} \emph{\bibinfo{volume}{38}}, \bibinfo{pages}{2232--2239}. \DOIprefix\doi{10.1364/AO.38.002232}.
\bibitem[{Krishnan et~al.(2011)Krishnan, Tay and Fergus}]{krishan11}
\bibinfo{author}{Krishnan, D.}, \bibinfo{author}{Tay, T.}, and \bibinfo{author}{Fergus, R.} (\bibinfo{year}{2011}). \bibinfo{title}{Blind deconvolution using a normalized sparsity measure}.
\newblock In \bibinfo{booktitle}{CVPR 2011}. pp. \bibinfo{pages}{233--240}.
\newblock \DOIprefix\doi{10.1109/CVPR.2011.5995521}.
\bibitem[{Trong et~al.(2014)Trong, Phuong, Tuyen and Thanh}]{Trong14}
\bibinfo{author}{Trong, D.D.}, \bibinfo{author}{Phuong, C.X.}, \bibinfo{author}{Tuyen, T.T.}, and \bibinfo{author}{Thanh, D.N.} (\bibinfo{year}{2014}). \bibinfo{title}{Tikhonov’s regularization to the deconvolution problem}.
\newblock \bibinfo{journal}{Communications in Statistics - Theory and Methods} \emph{\bibinfo{volume}{43}}, \bibinfo{pages}{4384–4400}. \DOIprefix\doi{10.1080/03610926.2012.721916}.
\bibitem[{Brody(2007)}]{Brody07}
\bibinfo{author}{Brody, D.C.} (\bibinfo{year}{2007}). \bibinfo{title}{A note on exponential families of distributions}.
\newblock \bibinfo{journal}{Journal of Physics A: Mathematical and Theoretical} \emph{\bibinfo{volume}{40}}, \bibinfo{pages}{F691}. \DOIprefix\doi{10.1088/1751-8113/40/30/F01}.
\bibitem[{Munkhammar et~al.(2017)Munkhammar, Mattsson and Rydén}]{Munkhammar17}
\bibinfo{author}{Munkhammar, J.}, \bibinfo{author}{Mattsson, L.}, and \bibinfo{author}{Rydén, J.} (\bibinfo{year}{2017}). \bibinfo{title}{Polynomial probability distribution estimation using the method of moments}.
\newblock \bibinfo{journal}{PLOS ONE} \emph{\bibinfo{volume}{12}}, \bibinfo{pages}{1--14}. \DOIprefix\doi{10.1371/journal.pone.0174573}.
\bibitem[{Sriperumbudur et~al.(2017)Sriperumbudur, Fukumizu, Gretton, Hyv\"{a}rinen and Kumar}]{Sriperumbudur17}
\bibinfo{author}{Sriperumbudur, B.}, \bibinfo{author}{Fukumizu, K.}, \bibinfo{author}{Gretton, A.}, \bibinfo{author}{Hyv\"{a}rinen, A.}, and \bibinfo{author}{Kumar, R.} (\bibinfo{year}{2017}). \bibinfo{title}{Density estimation in infinite dimensional exponential families}.
\newblock \bibinfo{journal}{Journal of Machine Learning Research} \emph{\bibinfo{volume}{18}}, \bibinfo{pages}{1--59}.
\bibitem[{Pessoa et~al.(2021)Pessoa, Costa and Caticha}]{Pessoa21}
\bibinfo{author}{Pessoa, P.}, \bibinfo{author}{Costa, F.X.}, and \bibinfo{author}{Caticha, A.} (\bibinfo{year}{2021}). \bibinfo{title}{Entropic dynamics on {Gibbs} statistical manifolds}.
\newblock \bibinfo{journal}{Entropy} \emph{\bibinfo{volume}{23}}, \bibinfo{pages}{494}. \DOIprefix\doi{10.3390/e23050494}.
\bibitem[{Kilic et~al.(2021)Kilic, Sgouralis and Pressé}]{Kilic21}
\bibinfo{author}{Kilic, Z.}, \bibinfo{author}{Sgouralis, I.}, and \bibinfo{author}{Pressé, S.} (\bibinfo{year}{2021}). \bibinfo{title}{Generalizing hmms to continuous time for fast kinetics: Hidden markov jump processes}.
\newblock \bibinfo{journal}{Biophysical Journal} \emph{\bibinfo{volume}{120}}, \bibinfo{pages}{409–423}. \DOIprefix\doi{10.1016/j.bpj.2020.12.022}.
\bibitem[{Bryan~IV et~al.(2022)Bryan~IV, Sgouralis and Pressé}]{Bryan22}
\bibinfo{author}{Bryan~IV, J.S.}, \bibinfo{author}{Sgouralis, I.}, and \bibinfo{author}{Pressé, S.} (\bibinfo{year}{2022}). \bibinfo{title}{Diffraction-limited molecular cluster quantification with bayesian nonparametrics}.
\newblock \bibinfo{journal}{Nature Computational Science} \emph{\bibinfo{volume}{2}}, \bibinfo{pages}{102–111}. \DOIprefix\doi{10.1038/s43588-022-00197-1}.
\bibitem[{Sukys et~al.(2022)Sukys, Öcal and Grima}]{Sukys22}
\bibinfo{author}{Sukys, A.}, \bibinfo{author}{Öcal, K.}, and \bibinfo{author}{Grima, R.} (\bibinfo{year}{2022}). \bibinfo{title}{Approximating solutions of the chemical master equation using neural networks}.
\newblock \bibinfo{journal}{iScience} \emph{\bibinfo{volume}{25}}, \bibinfo{pages}{105010}. \DOIprefix\doi{https://doi.org/10.1016/j.isci.2022.105010}.
\bibitem[{Rezende and Mohamed(2015)}]{Rezende15}
\bibinfo{author}{Rezende, D.}, and \bibinfo{author}{Mohamed, S.} (\bibinfo{year}{2015}). \bibinfo{title}{Variational inference with normalizing flows}.
\newblock In \bibinfo{editor}{ F.{ }Bach}, and \bibinfo{editor}{ D.{ }Blei}, eds. \bibinfo{booktitle}{Proceedings of the 32nd International Conference on Machine Learning} vol.~\bibinfo{volume}{37} of \emph{\bibinfo{series}{Proceedings of Machine Learning Research}}. \bibinfo{address}{Lille, France}: \bibinfo{publisher}{PMLR} pp. \bibinfo{pages}{1530--1538}.
\bibitem[{Papamakarios et~al.(2017)Papamakarios, Pavlakou and Murray}]{Papamakarios17}
\bibinfo{author}{Papamakarios, G.}, \bibinfo{author}{Pavlakou, T.}, and \bibinfo{author}{Murray, I.} (\bibinfo{year}{2017}). \bibinfo{title}{Masked autoregressive flow for density estimation}.
\newblock In \bibinfo{editor}{ I.{ }Guyon}, \bibinfo{editor}{ U.V.{ }Luxburg}, \bibinfo{editor}{ S.{ }Bengio}, \bibinfo{editor}{ H.{ }Wallach}, \bibinfo{editor}{ R.{ }Fergus}, \bibinfo{editor}{ S.{ }Vishwanathan}, and \bibinfo{editor}{ R.{ }Garnett}, eds. \bibinfo{booktitle}{Advances in Neural Information Processing Systems} vol.~\bibinfo{volume}{30}. \bibinfo{publisher}{Curran Associates, Inc.}
\bibitem[{Durkan et~al.(2019)Durkan, Bekasov, Murray and Papamakarios}]{Durkan19}
\bibinfo{author}{Durkan, C.}, \bibinfo{author}{Bekasov, A.}, \bibinfo{author}{Murray, I.}, and \bibinfo{author}{Papamakarios, G.} (\bibinfo{year}{2019}). \bibinfo{title}{Neural spline flows}.
\newblock In \bibinfo{editor}{ H.{ }Wallach}, \bibinfo{editor}{ H.{ }Larochelle}, \bibinfo{editor}{ A.{ }Beygelzimer}, \bibinfo{editor}{d\textquotesingle  F.{ }Alch\'{e}-Buc}, \bibinfo{editor}{ E.{ }Fox}, and \bibinfo{editor}{ R.{ }Garnett}, eds. \bibinfo{booktitle}{Advances in Neural Information Processing Systems} vol.~\bibinfo{volume}{32}. \bibinfo{publisher}{Curran Associates, Inc.}
\bibitem[{Dockhorn et~al.(2020)Dockhorn, Ritchie, Yu and Murray}]{Dockhorn20}
\bibinfo{author}{Dockhorn, T.}, \bibinfo{author}{Ritchie, J.A.}, \bibinfo{author}{Yu, Y.}, and \bibinfo{author}{Murray, I.} (\bibinfo{year}{2020}). \bibinfo{title}{Density deconvolution with normalizing flows}.
\newblock \bibinfo{journal}{ArXiv} \emph{\bibinfo{volume}{abs/2006.09396}}.
\bibitem[{Kobyzev et~al.(2021)Kobyzev, Prince and Brubaker}]{Kobyzev21}
\bibinfo{author}{Kobyzev, I.}, \bibinfo{author}{Prince, S.J.}, and \bibinfo{author}{Brubaker, M.A.} (\bibinfo{year}{2021}). \bibinfo{title}{Normalizing flows: An introduction and review of current methods}.
\newblock \bibinfo{journal}{IEEE Transactions on Pattern Analysis and Machine Intelligence} \emph{\bibinfo{volume}{43}}, \bibinfo{pages}{3964–3979}. \DOIprefix\doi{10.1109/tpami.2020.2992934}.
\bibitem[{Stimper et~al.(2023)Stimper, Liu, Campbell, Berenz, Ryll, Schölkopf and Hernández-Lobato}]{Stimper23}
\bibinfo{author}{Stimper, V.}, \bibinfo{author}{Liu, D.}, \bibinfo{author}{Campbell, A.}, \bibinfo{author}{Berenz, V.}, \bibinfo{author}{Ryll, L.}, \bibinfo{author}{Schölkopf, B.}, and \bibinfo{author}{Hernández-Lobato, J.M.} (\bibinfo{year}{2023}). \bibinfo{title}{normflows: A {PyTorch} package for normalizing flows}.
\newblock \bibinfo{journal}{Journal of Open Source Software} \emph{\bibinfo{volume}{8}}, \bibinfo{pages}{5361}. \DOIprefix\doi{10.21105/joss.05361}.
\bibitem[{Goodfellow et~al.(2016)Goodfellow, Bengio and Courville}]{Goodfellow16}
\bibinfo{author}{Goodfellow, I.}, \bibinfo{author}{Bengio, Y.}, and \bibinfo{author}{Courville, A.} (\bibinfo{year}{2016}). \bibinfo{title}{Deep Learning}. \bibinfo{publisher}{MIT Press}.
\newblock \bibinfo{note}{\url{http://www.deeplearningbook.org}}.
\bibitem[{Pessoa(2025)}]{github}
\bibinfo{author}{Pessoa, P.} (\bibinfo{year}{2025}).
\newblock \bibinfo{title}{{NFdeconvolve}}.
\newblock \bibinfo{howpublished}{\url{https://github.com/PessoaP/NFdeconvolve }}. .
\bibitem[{Paszke et~al.(2019)Paszke, Gross, Massa, Lerer, Bradbury, Chanan, Killeen, Lin, Gimelshein, Antiga, Desmaison, Kopf, Yang, DeVito, Raison, Tejani, Chilamkurthy, Steiner, Fang, Bai and Chintala}]{pytorch}
\bibinfo{author}{Paszke, A.}, \bibinfo{author}{Gross, S.}, \bibinfo{author}{Massa, F.}, \bibinfo{author}{Lerer, A.}, \bibinfo{author}{Bradbury, J.}, \bibinfo{author}{Chanan, G.}, \bibinfo{author}{Killeen, T.}, \bibinfo{author}{Lin, Z.}, \bibinfo{author}{Gimelshein, N.}, \bibinfo{author}{Antiga, L.}, \bibinfo{author}{Desmaison, A.}, \bibinfo{author}{Kopf, A.}, \bibinfo{author}{Yang, E.}, \bibinfo{author}{DeVito, Z.}, \bibinfo{author}{Raison, M.}, \bibinfo{author}{Tejani, A.}, \bibinfo{author}{Chilamkurthy, S.}, \bibinfo{author}{Steiner, B.}, \bibinfo{author}{Fang, L.}, \bibinfo{author}{Bai, J.}, and \bibinfo{author}{Chintala, S.} (\bibinfo{year}{2019}). \bibinfo{title}{{PyTorch: An Imperative Style, High-Performance Deep Learning Library}}.
\newblock In \bibinfo{editor}{ H.{ }Wallach}, \bibinfo{editor}{ H.{ }Larochelle}, \bibinfo{editor}{ A.{ }Beygelzimer}, \bibinfo{editor}{d\textquotesingle  F.{ }Alch\'{e}-Buc}, \bibinfo{editor}{ E.{ }Fox}, and \bibinfo{editor}{ R.{ }Garnett}, eds. \bibinfo{booktitle}{Advances in Neural Information Processing Systems 32} pp. \bibinfo{pages}{8024--8035}.. \bibinfo{publisher}{Curran Associates, Inc.} pp. \bibinfo{pages}{8024--8035}.
\bibitem[{Moulines et~al.(1997)Moulines, Cardoso and Gassiat}]{Moulines97}
\bibinfo{author}{Moulines, E.}, \bibinfo{author}{Cardoso, J.F.}, and \bibinfo{author}{Gassiat, E.} (\bibinfo{year}{1997}). \bibinfo{title}{Maximum likelihood for blind separation and deconvolution of noisy signals using mixture models}.
\newblock In \bibinfo{booktitle}{1997 IEEE International Conference on Acoustics, Speech, and Signal Processing} vol.~\bibinfo{volume}{5} of \emph{\bibinfo{series}{ICASSP-97}}. \bibinfo{publisher}{IEEE Comput. Soc. Press} pp. \bibinfo{pages}{3617–3620}.
\newblock \DOIprefix\doi{10.1109/icassp.1997.604649}.
\bibitem[{Santamaria et~al.(1999)Santamaria, Pantaleon, Ibanez and Artes}]{Santamaria99}
\bibinfo{author}{Santamaria, I.}, \bibinfo{author}{Pantaleon, C.}, \bibinfo{author}{Ibanez, J.}, and \bibinfo{author}{Artes, A.} (\bibinfo{year}{1999}). \bibinfo{title}{Deconvolution of seismic data using adaptive gaussian mixtures}.
\newblock \bibinfo{journal}{IEEE Transactions on Geoscience and Remote Sensing} \emph{\bibinfo{volume}{37}}, \bibinfo{pages}{855–859}. \DOIprefix\doi{10.1109/36.752203}.
\bibitem[{Bovy et~al.(2011)Bovy, Hogg and Roweis}]{Bovy11}
\bibinfo{author}{Bovy, J.}, \bibinfo{author}{Hogg, D.W.}, and \bibinfo{author}{Roweis, S.T.} (\bibinfo{year}{2011}). \bibinfo{title}{Extreme deconvolution: Inferring complete distribution functions from noisy, heterogeneous and incomplete observations}.
\newblock \bibinfo{journal}{The Annals of Applied Statistics} \emph{\bibinfo{volume}{5}}. \DOIprefix\doi{10.1214/10-aoas439}.
\bibitem[{Sorek et~al.(2024)Sorek, Haim, Chalifa-Caspi, Lazarescu, Ziv-Agam, Hagemann, Nono~Nankam, Bl\"{u}her, Liberty, Dukhno, Kukeev, Yeger-Lotem, Rudich and Levin}]{Sorek24}
\bibinfo{author}{Sorek, G.}, \bibinfo{author}{Haim, Y.}, \bibinfo{author}{Chalifa-Caspi, V.}, \bibinfo{author}{Lazarescu, O.}, \bibinfo{author}{Ziv-Agam, M.}, \bibinfo{author}{Hagemann, T.}, \bibinfo{author}{Nono~Nankam, P.A.}, \bibinfo{author}{Bl\"{u}her, M.}, \bibinfo{author}{Liberty, I.F.}, \bibinfo{author}{Dukhno, O.}, \bibinfo{author}{Kukeev, I.}, \bibinfo{author}{Yeger-Lotem, E.}, \bibinfo{author}{Rudich, A.}, and \bibinfo{author}{Levin, L.} (\bibinfo{year}{2024}). \bibinfo{title}{{sNucConv}: A bulk rna-seq deconvolution method trained on single-nucleus rna-seq data to estimate cell-type composition of human adipose tissues}.
\newblock \bibinfo{journal}{iScience} \emph{\bibinfo{volume}{27}}, \bibinfo{pages}{110368}. \DOIprefix\doi{10.1016/j.isci.2024.110368}.
\bibitem[{Sgouralis et~al.(2024)Sgouralis, Xu, Jalihal, Kilic, Walter and Pressé}]{Sgouralis24}
\bibinfo{author}{Sgouralis, I.}, \bibinfo{author}{Xu, L.W.Q.}, \bibinfo{author}{Jalihal, A.P.}, \bibinfo{author}{Kilic, Z.}, \bibinfo{author}{Walter, N.G.}, and \bibinfo{author}{Pressé, S.} (\bibinfo{year}{2024}). \bibinfo{title}{{BNP-Track: a framework for superresolved tracking}}.
\newblock \bibinfo{journal}{Nature Methods} \emph{\bibinfo{volume}{21}}, \bibinfo{pages}{1716–1724}. \DOIprefix\doi{10.1038/s41592-024-02349-9}.
\bibitem[{Teng et~al.(2024)Teng, Schaaf, Abulez, Hood, Wilson, Litzi, Mitchell, Conrads, Hunt, Olowu, Oliver, Park, Edwards, Chiang, Wilkerson, Raj-Kumar, Tarney, Darcy, Phippen, Maxwell, Conrads and Bateman}]{Teng24}
\bibinfo{author}{Teng, P.}, \bibinfo{author}{Schaaf, J.P.}, \bibinfo{author}{Abulez, T.}, \bibinfo{author}{Hood, B.L.}, \bibinfo{author}{Wilson, K.N.}, \bibinfo{author}{Litzi, T.J.}, \bibinfo{author}{Mitchell, D.}, \bibinfo{author}{Conrads, K.A.}, \bibinfo{author}{Hunt, A.L.}, \bibinfo{author}{Olowu, V.}, \bibinfo{author}{Oliver, J.}, \bibinfo{author}{Park, F.S.}, \bibinfo{author}{Edwards, M.}, \bibinfo{author}{Chiang, A.}, \bibinfo{author}{Wilkerson, M.D.}, \bibinfo{author}{Raj-Kumar, P.K.}, \bibinfo{author}{Tarney, C.M.}, \bibinfo{author}{Darcy, K.M.}, \bibinfo{author}{Phippen, N.T.}, \bibinfo{author}{Maxwell, G.L.}, \bibinfo{author}{Conrads, T.P.}, and \bibinfo{author}{Bateman, N.W.} (\bibinfo{year}{2024}). \bibinfo{title}{{ProteoMixture}: A cell type deconvolution tool for bulk tissue proteomic data}.
\newblock \bibinfo{journal}{iScience} \emph{\bibinfo{volume}{27}}, \bibinfo{pages}{109198}. \DOIprefix\doi{10.1016/j.isci.2024.109198}.

\end{thebibliography}

\newpage
\appendix
\section{Monte Carlo Methods}\label{ApSec:MC}

In this section, we overview the sampling strategy used to approximate the posterior distributions in both Bayesian strategies: $p(\theta_B|\{x\},\theta_A)$, defined in \eqref{Bayes} for the Bayesian model with known parameters, and $p(\Psi_B|\{x\},\theta_A)$, defined in \eqref{BayesGM} for the Bayesian model with Gaussian mixture methods.  

In short, these samplers are Markov Chain Monte Carlo (MCMC) realizations of each posterior. For the known model example, $p(\theta_B|\{x\},\theta_A)$, the sampler generates a sequence $\{\theta_B^1,\theta_B^2, \ldots, \theta_B^S\}$ such that we approximate the integral in \eqref{Bayes-Reconstruction} as:
\begin{equation}
    p(b|\{x\},\theta_A)  = \int \dd \theta_B \ p_B(b|\theta_B) \ p(\theta_B|\{x\},\theta_A)  \ \approx \frac{1}{S}\sum_{s=1}^S p_B(b|\theta_B^s).
\end{equation}
An equivalent expression holds for $\Psi_B$ in \eqref{Gaussmix-reconstruction}. Additionally, we approximate the MAP estimate, as defined in \eqref{Bayes-MAP} and \eqref{Gaussmix-MAP}, using the sampled value leading to the MAP value:
\begin{equation}
    \theta_B^\text{MAP} (\{x\},\theta_A) \approx \arg \max\limits_{s} p(\theta^s_B|\{x\},\theta_A).
\end{equation}
In the following subsections, we explain the methods used to construct these MCMC realizations and elaborate on the choice of the prior for the Gaussian mixture.

\subsection{MCMC for Bayesian Model with Known Parameters}\label{ApSec:MCsum}

\subsubsection{Sum Example}
For the example described in Sec. \ref{sec:sum}, we have parameters $\theta_B = \{\alpha_B, \beta_B\}$, where the distribution of $b$, $p(B|\theta_B)$, follows a gamma distribution as defined in \eqref{Gamma}, and $a$ corresponds to Gaussian noise with known mean and variance, $\theta_A = \{\mu_A = 10,\sigma^2_A=1\}$. The posterior can then be expressed as
\begin{equation}
    p(\theta_B|\{x\},\theta_A) \propto p(\{x\}|\theta_A,\theta_B) p(\theta_B) = p(\theta_B) \prod_{n=1}^N p_X(x_n|\theta_A,\theta_B).
\end{equation}

\textbf{ The prior} used in the results presented in the main text is a log-Gaussian prior, expressed as
\begin{equation}\label{SIprior}
    p(\theta_B) = \frac{1}{\alpha_B \sqrt{2\pi \xi^2}} \exp\left[-\frac{1}{2} \frac{(\log \alpha_B)^2}{\xi^2}\right] 
    \frac{1}{\beta_B \sqrt{2\pi \xi^2}} \exp\left[-\frac{1}{2} \frac{(\log \beta_B)^2}{\xi^2}\right],
\end{equation}
with $\xi=100$. This prior confines $\alpha_B$ and $\beta_B$ to be strictly positive, as required by the gamma distribution, while remaining broad enough not to impose restrictive constraints.

\textbf{ The Likelihood} is derived by substituting the form of $p(b|\theta_B)$ from \eqref{Gamma} into \eqref{sum-convolution} yielding
\begin{equation}
\begin{split}
    p(x_n|\theta_A, \theta_B) &= \int_{-\infty}^\infty \dd b \ p_A(x_n - b|\theta_A) \ p_B(b|\theta_B) \\
    &= \int_0^\infty \dd b \frac{1}{\sqrt{2\pi\sigma_A^2}} \exp \left[ - \frac{1}{2} \frac{(x_n-b-\mu_A)^2}{\sigma^2_A} \right]  \  \frac{\beta_B^{\alpha_B}}{\Gamma(\alpha_B)} b^{\alpha_B-1} e^{- \beta_B b}.
\end{split}
\end{equation}
the integration limit was changed to zero because the gamma distribution is only defined for positive argument, the probability is zero otherwise. 
Since, to the best of our knowledge, there is no closed-form expression exists for the above integral, we approximate it numerically as 
\begin{equation}
    p(x_n|\theta_A, \theta_B) = \Delta b \sum_{m=1}^M
    \frac{1}{\sqrt{2\pi\sigma_A^2}} \exp \left[ - \frac{1}{2} \frac{(x_n-b_m-\mu_A)^2}{\sigma^2_A} \right] 
    \frac{\beta_B^{\alpha_B}}{\Gamma(\alpha_B)} b_m^{\alpha_B-1} e^{- \beta_B b_m}.
\end{equation}
Where the $M=20000$ integration points,  $\{b_1, b_2, \ldots, b_M\}$, are equally spaced such that $b_{m+1} = b_m + \Delta b$, with $b_1=0.001$ and $b_M = \max\limits_n x_n - \mu_A/2$. 

\textbf{Initializing the Sampler} means determine the initial value $\theta_B^1$ by locating a region where the posterior density is high, reducing the need to discard early samples. Here we  obtained this initialization via a grid search:
\begin{equation}
    \theta_B^1 = \arg \max_{k, l} P(\theta_B = \{\alpha_l, \beta_k\}  | \{x\}, \theta_A),
\end{equation}
where $\{\alpha_l, \beta_k\}$ represent grid points for parameters $\alpha$ and $\beta$, with $\alpha_l$ equally spaced between $0.001$ and $10$, and $\beta_k$ equally spaced between $0.001$ and $6$.

\textbf{MCMC Proposal}  
With the initial value $\theta_B^1$, the sampler iteratively generates samples from the posterior distribution. At each iteration $s$, the sampler generates a new value value $\theta_B^{\text{prop}}$ using a proposal distribution $q(\theta_B^{\text{prop}}  | \theta_B^s)$. The next sample value, $\theta_B^{s+1}$, is determined as
\begin{equation}\label{MHdefine}
    \theta_B^{s+1} = 
    \begin{cases} 
        \theta_B^{\text{prop}} & \quad\text{if } u < A(\theta_B^{\text{prop}}, \theta_B^s) \ , \\
        \theta_B^s & \quad \text{otherwise}
    \end{cases}
\end{equation}
where $u $ is sampled uniformly between 0 and 1 and $A(\theta_B^{\text{prop}}, \theta_B^s)$ is the Metropolis-Hastings acceptance rate, 
\begin{equation}
    A(\theta_B^{\text{prop}}, \theta_B^s) = \min\left(1, \frac{p(\theta_B^{\text{prop}}  | \{x\}, \theta_A) }{p(\theta_B^s  | \{x\}, \theta_A)  | \theta_B^s)}\ \frac{q(\theta_B^s  | \theta_B^{\text{prop}})}{q(\theta_B^{\text{prop}}  | \theta_B^s)}\right).
\end{equation}
In other words, the proposal is accepted with probability $A(\theta_B^{\text{prop}}, \theta_B^s)$
For the sum example described in Sec \ref{sec:sum}, we use a proposal of the form of two log-Gaussian
\begin{equation}
    q(\theta_B^{\text{prop}} |\theta_B^s ) =  \frac{1}{\alpha_B \ \sqrt{2\pi \sigma_s^2} } \exp\left[-\frac{1}{2} \frac{\left(\log \frac{\alpha_B^{\text{prop}}}{\alpha_B^{s}}\right)^2}{\sigma_s^2}\right] 
    \ \frac{1}{\beta_B \ \sqrt{2\pi \sigma_s^2} } \exp\left[-\frac{1}{2} \frac{\left(\log \frac{\beta_B^{\text{prop}}}{\beta_B^{s}}\right)^2}{\sigma_s^2}\right]
\end{equation}
with $\sigma_s = .01$.  
The results presented in the main text are based on MCMC chains of $S=10000$ samples.

\subsubsection{Product example}
For the Bayesian with known model in the product example (Sec.\ref{sec:product})  we have the same parameters $\theta_b = \{ \alpha_B, \beta_B\}$ with the distribution of $b$,  is a gamma distribution and  Gaussian noise, with known mean and variance  $\theta_A = \{\mu_A = 10,\sigma^2_A=1\}$. As such, we can use the same prior in \eqref{SIprior}. 

\textbf{The likelihood}, however, will be calculated differently. Since we do in log space meaning we will use that that $\log x = \log a + \log b$, such that we can do the convolution as in \eqref{sum-convolution} such that 
\begin{equation}\label{conv-log}
\begin{split}
    p( x|\theta_A,\theta_B) 
& =  \left| \dv{\log x}{x} \right|  p(\log x|\theta_A,\theta_B) =  \frac{1}{|x|} p(\log x|\theta_A,\theta_B) \\
&=  \frac{1}{|x|} \int \dd (\log b) \ \   p(\log x - \log b|\theta_A) \ p(\log b|\theta_B) \\
&=  \frac{1}{|x|} \int \dd (\log b) \ \  \frac{p_A\left( \frac{x}{b}|\theta_A\right) }{ \left|\dv{\log\left(\frac{x}{b}\right)}{b} \right|} \frac{ p_B(b|\theta_B )} { \left| \dv{\log b}{b} \right|} \\
&=  \frac{1}{|x|} \int \dd (\log b) \ \ {p_A\left( \frac{x}{b}|\theta_A\right) } {\left|\frac{x}{b}\right|} \ { p_B(b|\theta_B )} { |b|} \\
&=   \int \dd (\log b) \ \ {p_A\left( \frac{x}{b}|\theta_A\right) } \ { p_B(b|\theta_B )} 
    \end{split}
\end{equation}
Note that this is equivalent to the Mellin convolution \eqref{prod-convolution}, as we could change variables the integral as
\begin{equation}
\begin{split}
    p( x|\theta_A,\theta_B) 
&=  \frac{1}{|x|} \int \dd b \ \left| \dv{\log b}{b} \right| \ \ {p_A\left( \frac{x}{b}|\theta_A\right) }  { p_B(b|\theta_B )}  \\
&=  \int \dd b \ { p_B(b|\theta_B )}   { p_A\left( \frac{x}{b}|\theta_A\right) } \  \  \frac{1}{|b|} .\\
    \end{split}
\end{equation}
Although both ways are equivalent, we implemented a numerical approximation of \eqref{conv-log} as
\begin{equation}
    p(x_n|\theta_A, \theta_B) =\Delta \phi^b \sum_{m=1}^M
    \frac{1}{\sqrt{2\pi\sigma_A^2}} \exp \left[ - \frac{1}{2} \frac{(x_n-e^{\phi^b_m}-\mu_A)^2}{\sigma^2_A} \right] 
    \frac{\beta_B^{\alpha_B}}{\Gamma(\alpha_B)} \left(e^{\phi^b_m}\right)^{\alpha_B-1} e^{- \beta_B e^{\phi^b_m}}.
\end{equation}
Where the $M=20000$ integration points,  $\{\phi^b_1, \phi^b_2, \ldots, \phi^b_M\}$ represent the range of $\log b$ and they are equally spaced such that $\phi^b_{m+1} = \phi^b_m + \Delta \phi^b$, with $\phi^b_1 = \min\limits_n \log x_n - \langle \log a \rangle - 3\sigma_{\log a}$ and $\phi^b_M= \max\limits_n x_n +3 \sigma_{\log a}. $
The MCMC initialization and proposals are otherwise like in the sum example.

\subsection{MCMC for Bayesian with Gaussian mixture}
\subsubsection{Prior}\label{ApSec:GMprior}
In the Bayesian with Gaussian mixture model, as written \eqref{infGaussian}, we assign distinct prior distributions to each set of parameters in $\Psi_B = \{ \bar{\mu}, \bar{\sigma^2}, \bar{\rho} \}$, representing $\bar{\mu}$, variances $\bar{\sigma^2}$, and weights $\bar{\rho}$. While the mixture model in \eqref{infGaussian} is formally expressed as an infinite sum, in practice truncate the sum at a finite upper limit $I$, we used $I=20$, resulting in a finite approximation where $\Psi_B$ is written as a array with $3I$ float elements.

For the results presented in the main text we a prior that is independent on each of these,
\begin{equation}
p(\Psi_B) = p(\bar{\mu}) p(\bar{\sigma^2}) p(\bar{\rho}) \ .
\end{equation}  
Where the factors in the equation above are, respectively: a Gaussian Prior over the Means ($\bar{\mu}$)
\begin{equation}
p(\bar{\mu}) = \prod_{i=1}^{I} \frac{1}{\sqrt{2 \pi \tau^2}} \exp\left[-\frac{(\mu_i - \xi)^2}{2 \tau^2}\right],
\end{equation}  
where $\xi=0$ is the prior mean and $\tau^ = 50$; a log-Gaussian prior for the variances
\begin{equation}
p(\bar{\sigma^2}) 
= \prod_{i=1}^{I} \frac{1}{\sigma_i^2 \zeta \sqrt{2 \pi}} \exp\left[-\frac{\left(\log(\sigma_i^2) - \eta \right)^2}{2 \zeta^2}\right],
\end{equation}  
with $\eta=0$ and $\zeta^2=1$; and, finally,  a Dirichlet prior over the weights 
\begin{equation}
p(\bar{\rho}) 
= \frac{\Gamma\left(\sum\limits_{i=1}^{I} \alpha_i\right)}{\prod\limits_{i=1}^{I} \Gamma(\alpha_i)} \prod\limits_{i=1}^{I} \rho_i^{\alpha_i - 1},
\end{equation}  
where $\alpha_i$ are the concentration parameters, we used
\begin{equation}
    \alpha_i = 10  \frac{(0.9)^i}{\sum\limits_{j=1}^I (0.9)^j } \ .
\end{equation}
In particular, the Dirichlet prior for $\bar{\rho}$ prevents overfitting by favoring decaying values of $\rho$ such that the first components are the most relevant, while ensuring that the weights respect the probability constrain $\sum_{i=1}^{I} \rho_i = 1$.

\subsubsection{Sum example}  
For the example described in Sec.  \ref{sec:sum}, the Gaussian mixture model is convoluted with the distribution $p_a$, which corresponds to Gaussian noise with known mean and variance, $\theta_A = \{\mu_A = 10, \sigma^2_A = 1\}$.

\textbf{The Likelihood}  
is straightforward because, interestingly, the convolution of two Gaussian distributions results in another Gaussian distribution. Furthermore, each component of the mixture is convoluted independently with the noise distribution $p_A$. As a result, the likelihood maintains the structure of a Gaussian mixture model with means shifted by $\mu_A$ and variances increased by $\sigma_A^2$:  
\begin{equation}\label{infGaussian2}
p(x | \Psi_B, \theta_B) = \sum_{i=1}^I \rho_i \frac{1}{\sqrt{2 \pi (\sigma_i^2 + \sigma_A^2)}} \exp\left[- \frac{(b - (\mu_i + \mu_A))^2}{2 (\sigma_i^2 + \sigma_A^2)} \right].
\end{equation}

\textbf{To initialize the sampler,}  
we used an initial state where all means $\mu_i$ were set equal to the data average shifted by $\mu_A$, $\mu_i = \langle x \rangle - \mu_A$, the variances $\sigma_i^2$ were initialized to the sample variance, $\sigma_i^2 = \mathrm{Var}(x) = \frac{1}{N} \sum_{n=1}^N (x_n - \overline{x})^2$, and the initial weights $\rho_i$ were set proportional to the Dirichlet prior concentration parameters $\alpha_i$: $\rho_i \propto \alpha_i$.  

Since this initialization creates a broad distribution and, as a consequence, does not guarantee that the parameters are in a region of large posterior probability, we optimized the full set of parameters $\Psi_B = \{ \bar{\mu}, \bar{\sigma^2}, \bar{\rho} \}$ using the PyTorch Adam optimizer. The negative log posterior was used as the loss function, and the optimization was run for 5000 steps before defining the initial state.

\textbf{Gibbs Sampler with Metropolis-Hastings Updates}  
After initializing the parameters $\Psi_B = \{ \bar{\mu}, \bar{\sigma^2}, \bar{\rho} \}$ using the Adam optimizer, we proceed with inference using a Gibbs sampling approach. In this scheme, we sequentially update one subset of parameters at a time ($\bar{\mu}$, $\bar{\sigma^2}$, or $\bar{\rho}$), from the current values of the other parameters.

Metropolis-Hastings (as seen in Sec.  \ref{ApSec:MCsum}) is used at each step with carefully chosen proposal distributions to ensure efficient sampling.

\textbf{To update the means}, $\bar{\mu}$, we use a Gaussian proposal distribution centered at the current sample element $\mu_i^{s}$:  
\begin{equation}
q_{\mu}(\mu_i^{\text{prop}} | \mu_i^{s}) = \frac{1}{\sqrt{2 \pi \tau_\mu^2}} \exp\left[-\frac{(\mu_i^{\text{prop}} - \mu_i^{s})^2}{2 \tau_\mu^2}\right],
\end{equation}  
where $\tau_\mu = 0.01$ is the proposal variance.  
Following Metropolis-Hastings, the proposed set of values $\bar{\mu}^{\text{prop}} = \{\mu_i^{\text{prop}}\}$ is accepted with probability:  
\begin{equation}
A (\bar{\mu}^{\text{prop}} | \bar{\mu}^{s}) = \min \left(1, \frac{p(\Psi_B^{\text{prop}} | \{x\},\theta_A ) }{p(\Psi_B^{s} | \{x\},\theta_A ) } \prod_{i=1}^I \frac{q_{\mu}(\mu_i^{\text{prop}} | \mu_i^{s})}{q_{\mu}(\mu_i^{s} | \mu_i^{\text{prop}})} \right),
\end{equation}  
where $\Psi_B^{\text{prop}} = (\bar{\mu}^{\text{prop}}, \bar{\sigma^2}^{s}, \bar{\rho}^{s})$.  
As in Sec.  \ref{MHdefine}, if the proposed set $\bar{\mu}^{\text{prop}}$ is accepted, the sample is updated as $\bar{\mu}^{s+1} = \bar{\mu}^{\text{prop}}$, otherwise, the previous sample is retained, $\bar{\mu}^{s+1} = \bar{\mu}^{s}$.  

\textbf{To update the variances}, $\bar{\sigma^2}$, we use a log-Gaussian proposal distribution centered at the logarithm of the current sample element $\log(\sigma_i^{2,s})$:  
\begin{equation}
q_{\sigma^2}(\sigma_i^{2,\text{prop}} | \sigma_i^{2,s}) = \frac{1}{\sigma_i^{2,\text{prop}} \tau_\sigma \sqrt{2 \pi}} \exp\left[-\frac{\left(\log(\sigma_i^{2,\text{prop}}) - \log(\sigma_i^{2,s})\right)^2}{2 \tau_\sigma^2}\right],
\end{equation}  
where $\tau_\sigma = 0.01$.  
Following Metropolis-Hastings, the proposed set of values $\bar{\sigma^2}^{\text{prop}} = \{\sigma_i^{2,\text{prop}}\}$ is accepted with probability:  
\begin{equation}
A (\bar{\sigma^2}^{\text{prop}} | \bar{\sigma^2}^{s}) = \min \left(1, \frac{p(\Psi_B^{\text{prop}} | \{x\},\theta_A ) }{p(\Psi_B^{s} | \{x\},\theta_A ) } \right),
\end{equation}  
where $\Psi_B^{\text{prop}} = (\bar{\mu}^{s+1}, \bar{\sigma^2}^{\text{prop}}, \bar{\rho}^{s})$.  

\textbf{To update the weights}, $\bar{\rho}$, we use a Dirichlet proposal distribution:  
\begin{equation}
q_{\rho}(\bar{\rho}^{\text{prop}} | \bar{\rho}^{s}) 
= \frac{\Gamma\left(\sum\limits_{i=1}^{I} \beta_i\right)}{\prod\limits_{i=1}^{I} \Gamma(\beta_i)} \prod\limits_{i=1}^{I} \rho_i^{\beta_i - 1},
\end{equation}
with concentration parameters $\beta_i = 10^5 \cdot \bar{\rho}_i^{s}$, which ensures a narrow proposal distribution around the previous sample. 
Following Metropolis-Hastings, the proposed set of values $\bar{\rho}^{\text{prop}}$ is accepted with probability
\begin{equation}
A (\bar{\rho}^{\text{prop}} | \rho^{s}) = \min \left(1, \frac{p(\Psi_B^{\text{prop}} | \{x\},\theta_A ) }{p(\Psi_B^{s} | \{x\},\theta_A ) } \right),
\end{equation}  
where $\Psi_B^{\text{prop}} = (\bar{\mu}^{s+1}, \bar{\sigma^2}^{\text{prop}}, \bar{\rho}^{s})$.  

The complete sample after one Gibbs iteration is then given by:  
\begin{equation}
\Psi_B^{s+1} = (\bar{\mu}^{s+1}, \bar{\sigma^2}^{s+1}, \bar{\rho}^{s+1}).
\end{equation}

After the initialization, we run a burn-in phase of 5000 samples of $\Psi_B$. These initial samples are discarded to avoid dependence on the starting conditions. The results presented in the main text are based on MCMC chains of $S = 20000$ samples, collected after this burn-in phase.

\subsubsection{Product example}  
To address the product example (Sec.  \ref{sec:product}) using a Gaussian mixture, we approximate the distribution of $\log b$ with a Gaussian mixture model. This transformation arises from rewriting the product $x = a b$ in logarithmic form as $\log x = \log a + \log b$. The Gaussian mixture model, parameterized by $\Psi_B = \{ \bar{\mu}, \bar{\sigma^2}, \bar{\rho} \}$, describes the probability distribution of $\log b$. This distribution of $a$, $p_A$, is a Gaussian of known mean and variance, given by $\theta_A = \{\mu_A = 10, \sigma^2_A = 1\}$.  

\textbf{The likelihood}, however, must be calculated numerically. Unlike the sum example, where the Gaussian nature of the sum was preserved, the distribution of $\log a$ is not Gaussian when the distribution of $a$ is Gaussian. Following an approach similar to Eq.  \eqref{conv-log}, we can write:  
\begin{equation}
\begin{split}
    p( x | \theta_A, \Psi_B) 
& =  \left| \dv{\log x}{x} \right|  p(\log x | \theta_A, \Psi_B) =  \frac{1}{|x|} p(\log x | \theta_A, \Psi_B) \\
&=  \frac{1}{|x|} \int \dd (\log b) \ \   p(\log x - \log b | \theta_A) \ p(\log b | \Psi_B) \\
&= \frac{1}{|x|}  \int \dd (\log b) \ \  \frac{p_A\left( \frac{x}{b} | \theta_A \right)}{ \left|\dv{\log\left(\frac{x}{b}\right)}{b} \right|}  p(\log b | \Psi_B) \\
&=   \int \dd (\log b) \ \ {p_A\left( \frac{x}{b} | \theta_A \right)} {\frac{1}{|b|}} \ { p_B(\log b | \Psi_B )}.
\end{split}
\end{equation}

We numerically approximate this integral as  
\begin{equation}
    p(x_n | \theta_A, \Psi_B) = \Delta \phi^b \sum_{m=1}^M
    \frac{1}{\sqrt{2\pi\sigma_A^2}} \exp \left[ - \frac{1}{2} \frac{(x_n-e^{\phi^b_m}-\mu_A)^2}{\sigma^2_A} \right]  \frac{1}{e^{\phi_m^b}}  \left(
    \sum_{i=1}^{I} \rho_i \frac{1}{\sqrt{2\pi \sigma_i^2}} \exp \left[- \frac{(\phi^b_m - \mu_i)^2}{2 \sigma_i^2} \right] \right) \ , 
\end{equation}
  where  $M = 20000$ is the number of integration points,  $\{\phi^b_1, \phi^b_2, \ldots, \phi^b_M\}$ represents the range of $\log b$, equally spaced such that $\phi^b_{m+1} = \phi^b_m + \Delta \phi^b$, and the range is defined as
\begin{equation}
\phi^b_1 = \min_n \log x_n - \langle \log a \rangle - 3 \sigma_{\log a}, \quad \phi^b_M = \max_n \log x_n + 3 \sigma_{\log a}.
\end{equation}

The Gaussian mixture model defined by $\Psi_B$ approximates the distribution of $\log b$, from which we obtain:  
\begin{equation}
    p_{B,\text{prod}}(b | \Psi_B) = \frac{1}{|b|}   \sum_{i=1}^{I} \rho_i \frac{1}{\sqrt{2\pi \sigma_i^2}} \exp \left[- \frac{(\log b - \mu_i)^2}{2 \sigma_i^2} \right].
\end{equation}

The MCMC initialization and proposal distributions follow the same structure described in the sum example. After initialization, we run a burn-in phase of 5000 samples of $\Psi_B$. These initial samples are discarded to eliminate dependence on the starting conditions. The results presented in the main text are based on MCMC chains of $S = 20000$ samples, collected after this burn-in phase.

\section{How \emph{NFdeconvolve} is trained}\label{ApSec:NFtraining}

In this section, we describe how \emph{NFdeconvolve} is trained. In particular, the most important part is how to calculate the likelihood through the convolution integral in \eqref{cov-nf} for the sum example ($x = a + b$). Later, we comment on how to generalize this to the product example ($x = ab$).

As described in Sec.\ref{sec:nfdeconvolve}, normalizing flows work by transforming a variable $z$ through a bijective transformation $f_\phi(z)$. This transformation is applied to an initial probability density $p_Z(z)$, resulting in the transformed density $p(y)$. Using the change of variable formula, this is expressed as:
\begin{equation}
p_{NF}(b|\phi) = p_Z(z) \left| \frac{\partial f_\phi}{\partial z} \right|^{-1}\ ,
\end{equation}
where $z = f_\phi^{-1}(b)$, making the equation above equivalent to \eqref{p_nf-def}. 
While normalizing flows can be viewed as a single transformation $f_\phi$, in practice, this is implemented as a sequence of composed bijective transformations:
\begin{equation}
f_\phi = f^L_{\phi_L} \circ f^{L-1}_{\phi_{L-1}} \circ \dots \circ f^1_{\phi_1} \ ,
\end{equation}
such that each $f^l_{\phi_l}$ is the $l$-th neural network layer with internal parameters $\phi_l$ and  $\phi$ represents the full set of parameters, including the parameters of all individual transformations $\phi =\{\phi_1, \phi_2, \dots, \phi_L \}$.

In NFdeconvole, we choose to have to have a standard Gaussian as the distribution for $z$ and initialize all $\phi_l$ such that all of the internal function are identity, as  default in the package used \cite{Stimper23}. However, to facilitate training we make so that the final layer as selected by hand to be an affine function $f^L_{\phi_L}(y) = \alpha  + \beta y$. Note that it makes so that $p_\text{NF}$ initializes as a Gaussian of center $\alpha$ and variance $\beta^2$. 

In particular, we chose the values for $\alpha$ and $\beta$ as
\begin{subequations} \label{alpha-beta} 
\begin{align} 
\alpha &= \overline{x} - \langle a \rangle \ , \\ 
\beta^2 &= \mathrm{Var}[x] \cdot \max\left(1 - \frac{\sigma_a^2}{\mathrm{Var}(x)}, \frac{1}{16} \right)\ ;
\end{align} 
\end{subequations}
where, $\overline{x}$ represents the sample mean of the x observations, $\overline{x} = \frac{1}{N} \sum_{n=1}^N x_n$ , an empirical estimate of the underlying mean while $\langle a \rangle$ denotes the expected value, calculates theoretically from $p_A$. Similarly, $\mathrm{Var}[x]$ is the sample variance,$\mathrm{Var}(x) = \frac{1}{N} \sum_{n=1}^N (x_n - \overline{x})^2$, while $\sigma_a^2$ refers to the variance of the underlying probability distribution $p_A$. In practice, NFdeconvolve receives $p_A$ as an object of PyTorch's distribution object\cite{pytorch} from which it automatically obtains the expected value $\overline{x}$ and variance $\sigma_a$.   
The values of $\alpha$ and $\beta$ are selected as in \eqref{alpha-beta} because if $x = a + b$, with $a$ and $b$ are independent,  it follows that $\langle b \rangle = \langle x \rangle - \langle a \rangle$ and  $\sigma_b^2 = \sigma_x^2 - \sigma_a^2$.  If we use $\overline{x} $ as an estimator for $\langle x \rangle$, and $\mathrm{Var}[x] $ as an estimator for $\sigma_x$, s a consequence, the normalizing flow is initialized as a Gaussian distribution with a good estimate for the mean and variance of $b$.  

With this initialization, at each training step, we aim to find $\phi^\ast$ as defined in \eqref{NF} by calculating the integral in \eqref{cov-nf} and using the optimization tools within PyTorch to find the maximum argument $\phi^\ast$. For the case where $x = a + b$, we could calculate the integral in \eqref{cov-nf} through a rectangle approximation as  
\begin{equation}\label{integrate}
\begin{split}
    p(x|\phi,\theta_A) &= \int_{-\infty}^{\infty} db \, p_{NF}(b|\phi) \, p_A(x - b|\theta_A) \\
    &\approx \Delta b \sum_{\mu=1}^M \, p_{NF}(b_\mu|\phi) \, p_A(x - b_\mu|\theta_A),
\end{split}
\end{equation}
where the integration points $\{b_1, b_2, \ldots, b_M\}$ are equally spaced such that $b_{\mu+1} = b_\mu + \Delta b$ for all $\mu$,  
\begin{equation}\label{int-limits}
b_1 = \min\limits_n x_n - \langle a \rangle - 3\sigma_a \quad \text{and} \quad b_M = \max\limits_n x_n - \langle a \rangle + 3\sigma_a.
\end{equation}
This range is chosen because we expect the probability of $b$ to be effectively zero beyond three standard deviations from the naively estimated values. The number of integration points, $M$, is set by default to 2,000 but can be adjusted by the user.

The training goal is to find 
\begin{equation}
\phi^\ast (\{x\},\theta_A) = \arg \max\limits_{\phi} \prod_{n=1}^N p(x_n|\phi,\theta_A),
\end{equation}
as defined in \eqref{NF}. Since probability densities are non-negative and the logarithm is a monotonic function, this can equivalently be written as 
\begin{equation}
    \phi^\ast (\{x\},\theta_A) = \arg \min\limits_{\phi} \mathcal{L}(\phi), \quad \text{where} \quad \mathcal{L}(\phi) =  - \frac{1}{N} \sum_{n=1}^N \log  p(x_n|\phi,\theta_A).
\end{equation}
In other words, we use $\mathcal{L}(\phi)$ as the loss function for the minimization problem, which can also be expressed, by substituting \eqref{integrate}, as 
\begin{equation} 
    \mathcal{L}(\phi) =  - \frac{1}{N} \sum_{n=1}^N  \text{lse}_{\substack{\mu}} 
    \Biggl( \log \Delta b + \log p_{NF}(b_\mu|\phi) + \log p_A(x_n - b_\mu|\theta_A) \Biggr),
\end{equation}
where $\text{lse}_\mu(x_\mu) = \log \big( \sum_\mu e^{x_\mu} \big)$. This formulation is preferred because the logarithm calculation is more numerically stable, and PyTorch packages are typically designed to work with logarithm of  probabilities.   

In the case where the observations are a product, $x = ab$, \emph{NFdeconvolve} performs calculations in log space, as described in Sec. \ref{sec:product}. The problem is transformed from $x = ab$ to $\log x = \log a + \log b$.  

\emph{NFdeconvolve} takes as input the observations $\{x\} = \{x_1, x_2, \ldots, x_N\}$ and the distribution of $a$, $p_A(a|\theta_A)$. Internally, these are transformed by constructing the effective dataset $\{\log x\} = \{ \log (x_1), \log (x_2),  \ldots,  \log (x_N)\}$ and converting the probability distribution of $a$ into $\log a$ as $ p(\log a|\theta_A) = p(a|\theta_A)  a$. 

When transforming the probability distribution, rough approximations of the expected and variance of $\log a$ are made via sampling. Specifically, $10000$ samples are drawn from $p_A$, \emph{NFdeconvolve}  takes  the logarithm of those from which the mean and sample variance are calculated. This approach is sufficient because only the expected value and variance of the distribution of $a$ are used to initialize the network training \eqref{alpha-beta} and set the (already conservative) integration limits \eqref{int-limits}.  The neural network training will refine these initial approximations.

Using the process described in this SI section we obtain the approximation of the distribution of $\log b$, $p_{NF}(\log b|\phi)$ from which we obtain the probability of b as $p_{NF}(b|\phi) = \frac{1}{|b|} p_{NF}(\log b|\theta_B)$. 

\end{document}